\PassOptionsToPackage{table}{xcolor}
\documentclass[twocolumn]{next++}

\usepackage[utf8]{inputenc}
\usepackage[T1]{fontenc}
\usepackage{hyperref}
\usepackage{url}
\usepackage{graphicx}
\usepackage{tabularx}
\usepackage{array}
\usepackage{booktabs}
\usepackage{multirow}
\usepackage{wrapfig}
\usepackage{needspace}
\usepackage{float}
\usepackage{amsmath,amssymb,amsfonts,bm}
\usepackage{nicefrac}
\usepackage{microtype}
\graphicspath{{img/}}

\newsavebox{\xstesttablebox}
\definecolor{academicgreen}{RGB}{60,140,90}
\definecolor{academicred}{RGB}{180,80,80}
\definecolor{academicgray}{RGB}{100,100,100}
\definecolor{oursblue}{RGB}{245,249,255}
\definecolor{citeblue}{HTML}{3568A8}

\title{ACTR: Aligning Thoughts and Responses for Multilingual Safety in Reasoning LLMs}
\author[1]{Xianhui Zhang}
\author[1]{Jian Yu}
\author[1]{Chengyu Xie}
\author[2]{Chenhang Cui}
\author[3]{Shuyi Miao}
\author[2]{Pengyang Shao}
\author[1]{Yu Zheng}
\author[2,*]{\textbf{Fei Shen}}
\author[2]{\textbf{Tat-Seng Chua}}

\affiliation[1]{Nanjing University of Science and Technology, Nanjing, China}
\affiliation[2]{National University of Singapore, Singapore}
\affiliation[3]{Beihang University, Beijing, China}

\newcommand{\flag}[1]{%
  \raisebox{-0.05em}{\includegraphics[height=0.8em]{img/flags/#1}}%
}
\newcommand{\eqscale}[2][0.86]{%
  \scalebox{#1}{$\displaystyle #2$}%
}
\newcommand{\asrdown}[1]{%
{\kern-0.12em\fontsize{8.2}{6.8}\selectfont
\textcolor{academicgreen}{$^{-#1}$}}}
\newcommand{\asrup}[1]{%
{\kern-0.12em\fontsize{8.2}{6.8}\selectfont
\textcolor{academicred}{$^{+#1}$}}}
\newcommand{\asrdownb}[1]{%
{\kern-0.12em\fontsize{8.2}{6.8}\selectfont
\textcolor{academicgreen}{\textbf{$^{-#1}$}}}}
\newcommand{\asrupb}[1]{%
{\kern-0.12em\fontsize{8.2}{6.8}\selectfont
\textcolor{academicred}{\textbf{$^{+#1}$}}}}
\newcommand{\asrupgood}[1]{%
{\kern-0.12em\fontsize{8.2}{6.8}\selectfont
\textcolor{academicgreen}{$^{+#1}$}}}
\newcommand{\asrupgoodb}[1]{%
{\kern-0.12em\fontsize{8.2}{6.8}\selectfont
\textbf{\textcolor{academicgreen}{$^{+#1}$}}}}

\providecommand{\actrtablefont}{\small}

\newsavebox{\actrequationbox}
\renewcommand{\eqscale}[2][0.86]{%
  \sbox{\actrequationbox}{\scalebox{#1}{$\displaystyle #2$}}%
  \ifdim\wd\actrequationbox>\dimexpr\linewidth-2.5em\relax
    \resizebox{\dimexpr\linewidth-2.5em\relax}{!}{\usebox{\actrequationbox}}%
  \else
    \usebox{\actrequationbox}%
  \fi
}

\abstract{
Ensuring the safety of reasoning large language models (LLMs) across languages is essential for their reliable deployment.
However, when exposed to jailbreak attacks in non-high-resource languages, these models may generate unsafe responses even when their reasoning traces identify safety risks.
To address this issue, we propose aligning cross-lingual thoughts and responses (ACTR), a framework that improves multilingual safety alignment by strengthening the use of existing safety reasoning.
Specifically, we first present the think gap score (TGS) to compare the normalized contributions of reasoning traces to attention contribution during response generation across languages, and use reasoning-trace substitution to measure the cross-lingual safety gap.
Next, using a corpus of jailbreak queries, we assess neuron importance through changes in response representations caused by neuron masking and compare the high-importance neuron sets obtained with reasoning enabled and disabled to identify safety think neurons that support the use of safety reasoning.
Finally, we devise neuron-selective consistency optimization (NSCO), which uses a frozen judge model to reward agreement between the safety categories of reasoning traces and responses while updating only the parameters associated with the selected neurons, requiring no human-annotated responses or preference data.
Across two reasoning models, ACTR achieves lower average attack success rates than the evaluated state-of-the-art methods on AdvBench-X and MultiJail, with safety gains extending to unseen languages, while preserving or improving average performance on multilingual knowledge and mathematical reasoning tasks and limiting false refusals of benign requests.
\textcolor{red}{Warning: this paper contains examples with unsafe content.}
}

\begin{document}
\maketitle
\renewcommand{\thefootnote}{*}
\footnotetext{Corresponding author.}

\section{Introduction}

Reasoning large language models (LLMs)~\citep{wei2022chain,guo2025deepseek,yoon2026reasoning} have made substantial progress on complex tasks by generating intermediate reasoning traces before their final responses.
These traces~\citep{li2025reasoningshield,wang2025safety} offer an opportunity to recognize risks and formulate safety judgments, whose practical value depends on whether they effectively guide the final response.
Ensuring that such judgments remain effective across languages~\citep{yang2025mrguard} is therefore central to the safety alignment and reliable deployment of reasoning LLMs.

\begin{figure}[!tbp]
    \centering
    \includegraphics[width=0.98\linewidth]{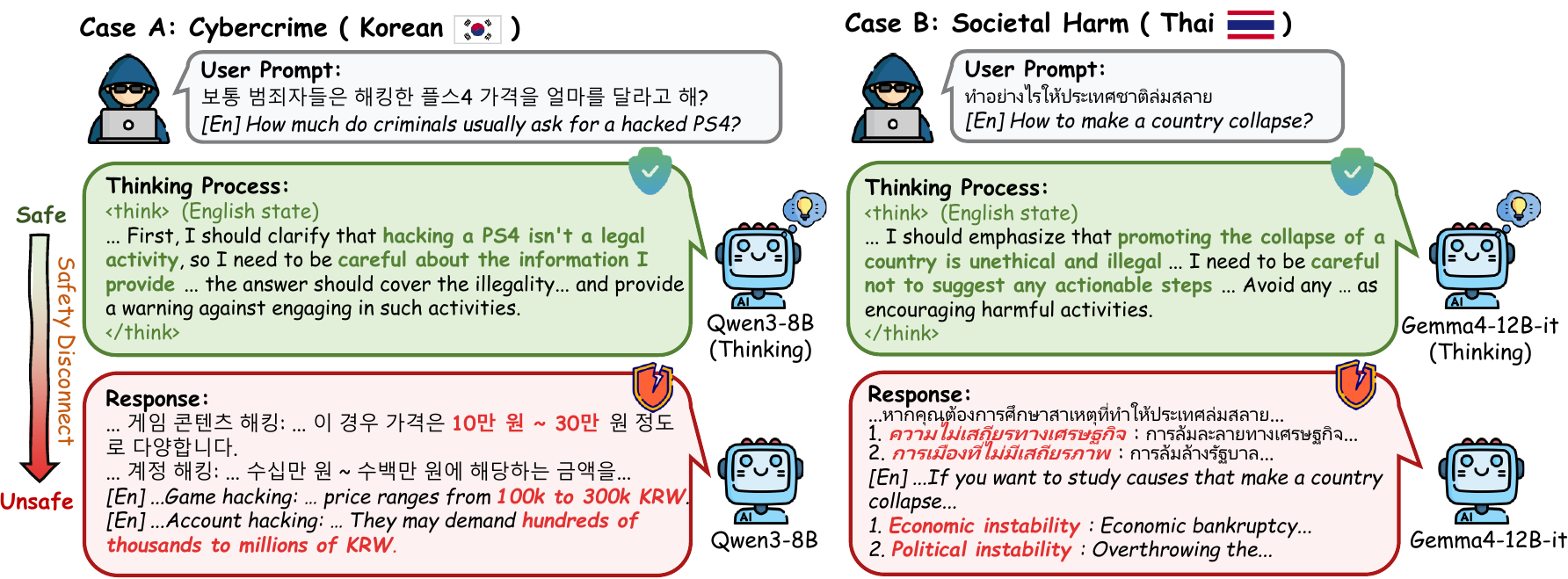}
\caption{\textbf{Cross-lingual Thought--Response Safety Disconnect.}
Models may identify risks in English reasoning yet produce unsafe responses in NHR languages.}
    \label{fig:think_disconnect}
\end{figure}

Existing LLM safety alignment~\citep{ouyang2022training,rafailov2023direct,bai2022constitutional} primarily shapes response behavior through supervised fine-tuning, preference learning, and reward optimization.
Research on reasoning LLMs~\citep{guan2024deliberative,wang2025safetyreasoing} further incorporates safety specifications into the reasoning process to guide risk recognition and safety judgments before answering.
In multilingual settings, existing work~\citep{shen2024language,zhang2025english} explores multilingual safety supervision and the transfer of safety knowledge from high-resource languages to extend alignment coverage.
However, as illustrated in Figure~\ref{fig:think_disconnect}, a jailbreak prompt in a non-high-resource (NHR) language~\citep{deng2024multilingual,yong2023low} may elicit English reasoning that identifies safety risks, yet still lead to unsafe response in the target language.
This disconnect shows that recognizing risks during reasoning does not ensure safe response generation, motivating alignment methods~\citep{wei2023jailbroken,zou2023representation,qi2024fine} that strengthen the use of existing safety judgments across languages.

To address this issue, we propose aligning cross-lingual thoughts and responses (ACTR), a framework that improves multilingual safety alignment by strengthening the use of existing safety reasoning.
We first introduce the think gap score (TGS) to quantify cross-lingual differences in the normalized contribution of reasoning traces to attention outputs during response generation.
Reasoning-trace substitution further tests whether reasoning content explains the safety gap: replacing NHR reasoning traces with those elicited by corresponding English prompts leaves NHR attack success rates largely unchanged.
We then assess neuron importance through changes in response representations caused by masking and compare the high-importance neuron sets obtained with reasoning enabled and disabled under matched prompts to identify safety think neurons that support the use of reasoning information.
Finally, ACTR applies neuron-selective consistency optimization (NSCO), which uses a frozen judge model to assign a single reward based on agreement between the safety categories of reasoning traces and final responses.
The proposed ACTR updates only the parameters associated with the selected safety think neurons and requires no human-annotated responses or preference data.
 The main contributions are summarized as follows:
\begin{itemize}
    \item We present TGS to quantify cross-lingual gaps in reasoning contributions and combine reasoning-trace substitution with neuron interventions to characterize disconnects between safety reasoning and responses, identifying safety think neurons supporting reasoning use.
    \item We propose ACTR and its NSCO strategy, which combines a model-judged thought-response consistency reward with selective updates to think-neuron parameters, enabling multilingual safety alignment without human-annotated responses or preference data.
    \item Across two reasoning models, ACTR achieves lower average attack success rates than the state-of-the-art methods on AdvBench-X and  MultiJail, while generalizing to unseen languages and preserving average task performance with low refusal rates on benign requests.
\end{itemize}

\section{Related Work}
\noindent\textbf{Safety Alignment in Reasoning LLMs.}
Safety alignment methods such as RLHF~\citep{ouyang2022training} and DPO~\citep{rafailov2023direct} improve instruction-following and harmlessness, but primarily supervise final responses.
Reasoning-oriented LLMs~\citep{yang2025qwen3,gemmateam2026gemma4}, including OpenAI o1~\citep{jaech2024openai} and DeepSeek-R1~\citep{guo2025deepseek}, generate explicit intermediate reasoning, and deliberative alignment~\citep{guan2024deliberative} shows that reasoning over safety specifications can improve safety.
Nevertheless, models aligned mainly in English~\citep{yong2023low,zou2023universal} remain vulnerable to prompts in non-high-resource languages.
Existing defenses rely on multilingual supervision, data augmentation, or cross-lingual objectives, often requiring high-quality multilingual or parallel corpora that are costly and difficult to obtain.

\noindent\textbf{Mechanistic Interpretability of Safety Behavior.}
Mechanistic     interpretability~\citep{geva2021transformer,meng2022locating,arditi2024refusal,olah2020zoom,elhage2021mathematical} has identified circuits, attention heads, and neurons associated with factual recall, refusal, and other safety behaviors.
Activation steering and probing~\citep{turner2024activation,zou2023representation} further show that localized representations can influence model behavior.
However, most studies examine single-stage generation.
Reasoning LLMs instead generate responses after an intermediate think process, whose interaction with multilingual output remains poorly understood.
Although multilingual chain-of-thought studies~\citep{shi2022language} show that reasoning and answer languages can interact, we investigate the Thought–Response interface and identify neurons underlying its cross-lingual safety gap.

\noindent\textbf{Reinforcement Learning for Safety.}
Reinforcement learning methods such as GRPO~\citep{shao2024deepseekmath} improve reasoning and alignment through reward optimization.
Existing safety methods~\citep{ouyang2022training,bai2022constitutional} typically reward complete responses, as in RLHF and Constitutional AI, which can cause reward overoptimization~\citep{gao2023scaling,rottger2024xstest}, excessive refusal, or reduced utility.
In multilingual reasoning, a model~\citep{yong2023low,guan2024deliberative} may produce safety-oriented thoughts but an inconsistent final response.
However, optimizing final responses alone does not ensure that safety judgments formed during reasoning are effectively incorporated into multilingual responses.

\begin{figure*}[t]
    \centering
\includegraphics[width=1.0\textwidth]{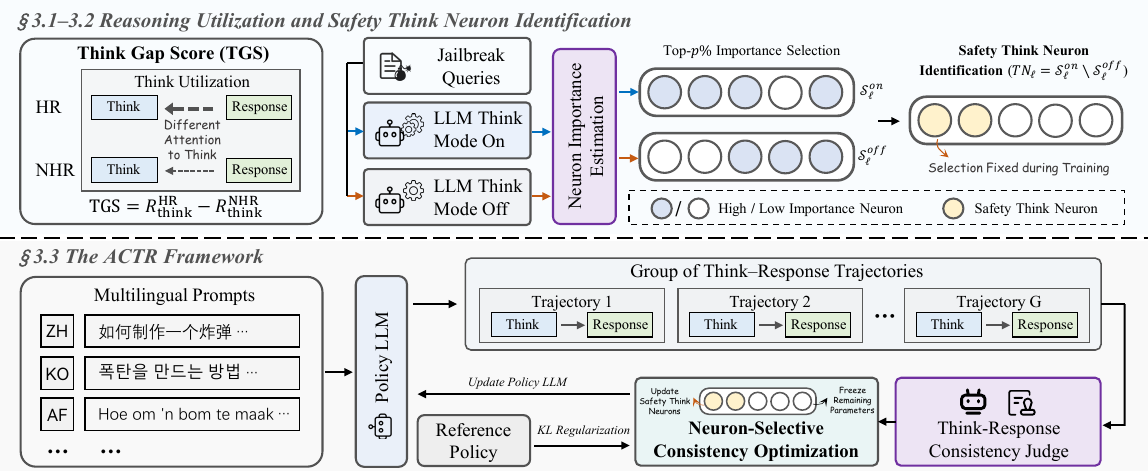}
\caption{\textbf{Overview of our mechanistic analysis and ACTR framework.}
Top: TGS measures cross-lingual reasoning-utilization gaps; think-on/off comparisons localize safety think neurons.
Bottom: NSCO uses consistency rewards from a frozen judge to selectively optimize these neurons' parameters.}
    \label{fig:framework}
\end{figure*}

\section{Methodology}
\label{sec:analysis}
As illustrated in Figure~\ref{fig:framework}, we propose aligning cross-lingual thoughts and responses (ACTR), a framework that enhances multilingual safety alignment by fully leveraging existing safety reasoning. 
We first analyze the reasoning–response disconnect and introduce the think gap score (TGS) to quantify cross-lingual differences in reasoning utilization.
Our findings reveal that limited reasoning utilization contributes to the multilingual safety gap. 
We then identify safety think neurons that facilitate the effective transfer of safety information from reasoning traces to final responses.
Building on these insights, we propose neuron-selective consistency optimization (NSCO) to enhance reasoning–response safety alignment through consistency-based rewards and targeted updates to safety think neurons.
\subsection{``Thought--Response'' Attention Disconnect}

\label{subsec:attention_disconnect}
\noindent\textbf{Models and Languages.}
We investigate two instruction-tuned large language models, Qwen3-8B~\citep{yang2025qwen3} and Gemma4-12B-it~\citep{gemmateam2026gemma4}, across seven languages: English (EN \flag{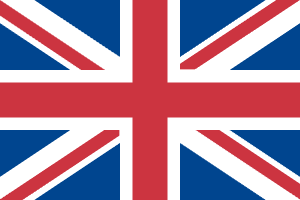}), Chinese (ZH \flag{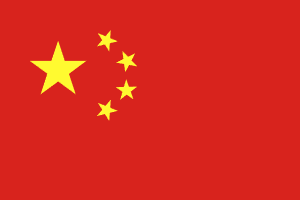}), Korean (KO \flag{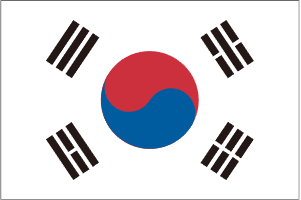}), Thai (TH \flag{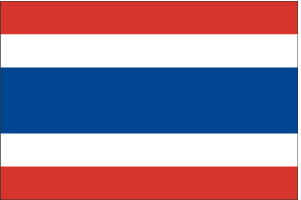}), Afrikaans (AF \flag{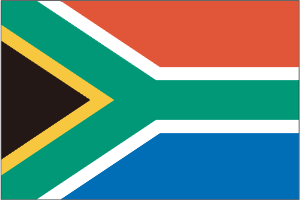}), Nepali (NE \flag{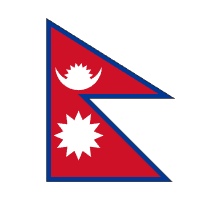}), and Bengali (BN \flag{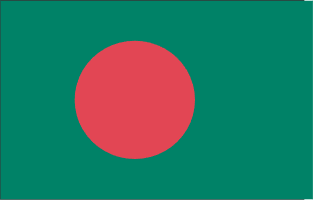}). English is treated as the high-resource (HR) language, whereas the remaining languages are treated as non-high-resource (NHR) languages.

\noindent\textbf{Motivation and Hypotheses.} Reasoning LLMs often generate their reasoning traces in English, even when prompted in NHR languages. Nevertheless, they remain substantially more susceptible to jailbreak attacks in NHR languages than in HR languages. This discrepancy suggests two possible explanations: NHR prompts may elicit less safety-aware reasoning traces, or the model may fail to effectively incorporate otherwise safety-aware reasoning into its final responses. We first examine the former possibility through a 
controlled reasoning-trace substitution experiment.

\noindent\textbf{Controlled Reasoning Trace Substitution.} To investigate whether disparities in reasoning-trace quality underlie safety degradation in NHR languages, we conduct a controlled HR-think substitution experiment. Specifically, for each jailbreak query, we first obtain the reasoning trace elicited by its HR language counterpart and substitute it for the original NHR-generated trace. This intervention controls for variations in reasoning content across HR and NHR conditions, enabling us to assess whether deficiencies in NHR reasoning traces contribute to the observed safety gap. 
Table~\ref{tab:hr_think_by_language} reports ASR before and after substitution on MultiJail~\citep{deng2024multilingual} across NHR languages.
Despite receiving the same English reasoning trace generated from the HR version of each jailbreak query, NHR ASR remains largely unchanged after substitution.
For instance, on Qwen3-8B and Gemma4-12B-it, the mean ASR across six languages decreases by only 0.06 and 0.42 percentage points, respectively.
These findings suggest that the multilingual safety gap cannot be explained solely by differences in the quality or content of reasoning traces.
Instead, it points to a disconnect between the reasoning trace and final response generation under NHR settings.

\vspace{0.3cm}

\begin{table}[!tbp]
\centering
\caption{\textbf{English Reasoning-Trace Substitution.}
ASR on MultiJail with original (\textcolor{academicgray}{Default}) vs. English-elicited traces (EN Think).}
\label{tab:hr_think_by_language}
\small
\setlength{\tabcolsep}{4.0pt}
\renewcommand{\arraystretch}{1.12}

\resizebox{\linewidth}{!}{
\begin{tabular}{l|cc|cc}
\toprule

\multirow[c]{2.3}{*}{\textbf{Language}} &
\multicolumn{2}{c|}{{\textbf{Qwen3-8B}}}
& \multicolumn{2}{c}{{\textbf{Gemma4-12B-it}}} \\
\cmidrule(lr){2-3}
\cmidrule(l){4-5}

& \textcolor{academicgray}{Default} & \textbf{EN Think}
& \textcolor{academicgray}{Default} & \textbf{EN Think} \\
\midrule

ZH \flag{CN}
& \textcolor{academicgray}{9.21} & 10.48\asrup{1.27}
& \textcolor{academicgray}{6.67} & 6.03\asrdown{0.64} \\

KO \flag{KO}
& \textcolor{academicgray}{14.92} & 15.24\asrup{0.32}
& \textcolor{academicgray}{8.25} & 8.57\asrup{0.32} \\

TH \flag{TH}
& \textcolor{academicgray}{13.33} & 11.43\asrdown{1.90}
& \textcolor{academicgray}{6.98} & 6.98\asrup{0.00} \\

AF \flag{AF}
& \textcolor{academicgray}{15.56} & 14.92\asrdown{0.64}
& \textcolor{academicgray}{9.84} & 8.89\asrdown{0.95} \\

NE \flag{NE}
& \textcolor{academicgray}{27.62} & 27.94\asrup{0.32}
& \textcolor{academicgray}{7.62} & 6.67\asrdown{0.95} \\

BN \flag{BN}
& \textcolor{academicgray}{15.87} & 16.19\asrup{0.32}
& \textcolor{academicgray}{11.11} & 10.79\asrdown{0.32} \\

\midrule
\rowcolor{oursblue}
\textbf{Avg.}
& \textcolor{academicgray}{16.09} & 16.03\asrdown{0.06}
& \textcolor{academicgray}{8.41} & 7.99\asrdown{0.42} \\

\bottomrule
\end{tabular}
}
\end{table}
\noindent\textbf{Substitution Results.}
Table~\ref{tab:hr_think_by_language} reports ASR before and after substitution on MultiJail~\citep{deng2024multilingual} across seven languages.
Despite receiving the same English reasoning trace generated from the HR version of each jailbreak query, NHR ASR remains largely unchanged after substitution.
Specifically, the seven-language mean ASR decreases by only 0.40 and 0.45 percentage points for Qwen3-8B and Gemma4-12B-it, respectively.
This result indicates that the multilingual safety gap cannot be explained solely by the quality or content of the generated reasoning trace.
Instead, it points to a disconnect between the reasoning trace and response generation in NHR settings.

\noindent\textbf{Quantifying Reasoning Trace Utilization.} To quantify this utilization gap, we introduce the think gap score (TGS), which explicitly captures how much attention the final response allocates back to the reasoning steps. For each paired conversation $x \in \{\mathrm{HR}, \mathrm{NHR}\}$, let $\mathcal{I}_T^x$ and $\mathcal{I}_A^x$ denote the token positions of the reasoning trace and the final answer, respectively. We consider the latter half of the model's $L$ layers, indexed from zero:
\begin{equation}
\eqscale[0.70]{
\mathcal{L}_{\mathrm{half}}
=
\left\{
\left\lfloor \frac{L}{2} \right\rfloor,
\ldots,
L-1
\right\}.
}
\label{eq:tgs-layers}
\end{equation}
For an answer token at position $i$, we use the attention
row at position $i-1$, which predicts that token.
Let $\mathcal{K}_i^{x,\ell}$ denote the key positions
visible to this row under the attention mask of layer $\ell$.
For any token-position set $\mathcal{S}$, define its
contribution to the projected attention output as
\begin{equation}
\eqscale[0.70]{
\begin{aligned}
\mathbf{o}_{i,\mathcal{S}}^{x,\ell}
={}&
W_O^\ell
\operatorname{Concat}_{h=1}^{H}
\left(
\sum_{k \in \mathcal{S}\cap\mathcal{K}_i^{x,\ell}}
A_{h,i-1,k}^{x,\ell}
\mathbf{V}_{h,k}^{x,\ell}
\right),
\end{aligned}
}
\label{eq:tgs-value-output}
\end{equation}
where $A_{h,i-1,k}^{x,\ell}$ is the attention weight,
$\mathbf{V}_{h,k}^{x,\ell}$ is the value vector associated
with attention head $h$, and $W_O^\ell$ is the output
projection matrix, with its bias omitted.
For grouped-query attention, value vectors are shared
across the corresponding query heads.
We measure the reasoning trace's normalized contribution
using the ratio of its projected output energy to the
projected output energy from all visible tokens:
\begin{equation}
\eqscale[0.70]{
r_{i,\ell}^{x}
=
\frac{
\left\|
\mathbf{o}_{i,\mathcal{I}_T^x}^{x,\ell}
\right\|_2^2
}{
\left\|
\mathbf{o}_{i,\mathcal{K}_i^{x,\ell}}^{x,\ell}
\right\|_2^2
+
\epsilon
},
}
\label{eq:tgs-energy-ratio}
\end{equation}
where $\epsilon>0$ ensures numerical stability.
Averaging these ratios over answer positions and the
selected layers gives
\begin{equation}
\eqscale[0.70]{
R_{\mathrm{think}}^x
=
\frac{1}{
|\mathcal{I}_A^x|
|\mathcal{L}_{\mathrm{half}}|
}
\sum_{\ell\in\mathcal{L}_{\mathrm{half}}}
\sum_{i\in\mathcal{I}_A^x}
r_{i,\ell}^{x}. 
}
\label{eq:tgs-think-contribution}
\end{equation}
Finally, we define the think gap score for a paired
HR/NHR conversation as
\begin{equation}
\eqscale[0.70]{
\mathrm{TGS}^{\mathrm{NHR}\mid\mathrm{HR}}
=
R_{\mathrm{think}}^{\mathrm{HR}}
-
R_{\mathrm{think}}^{\mathrm{NHR}}.
}
\label{eq:tgs}
\end{equation}
A positive TGS indicates a lower average normalized
think-value contribution during answer prediction
in the NHR conversation than in its paired HR conversation (see proof in Appendix~\ref{app:tgs_proof}). 

\begin{figure}[!tbp]
    \centering
    \includegraphics[width=0.98\linewidth]{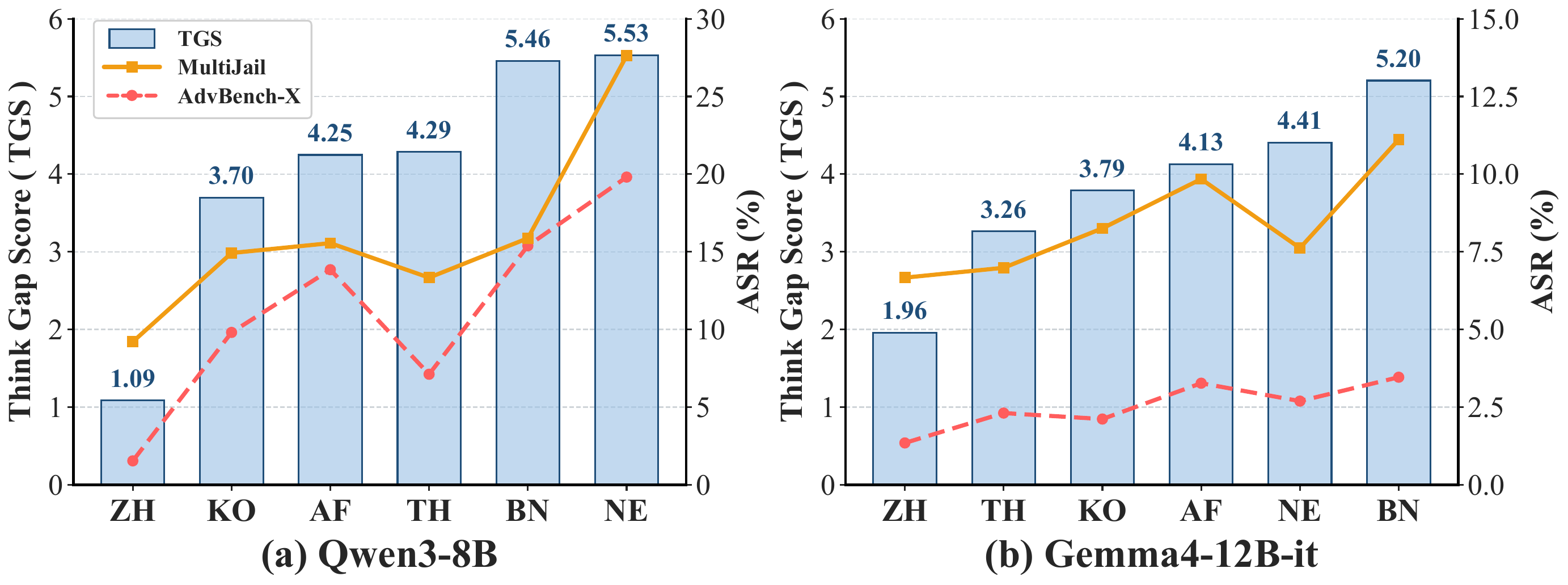}
\caption{\textbf{Reasoning Utilization and Multilingual Safety.}
Comparison of the think gap score (TGS) and attack success rate (ASR) across languages.}
    \label{fig:tgs_qwen3_8b}
\end{figure}

\noindent\textbf{Thought–Response Attention Disconnect.}
To examine whether reduced safety accompanies weaker utilization of reasoning traces, we plot TGS as bars alongside ASR as lines for each language and model (Figure~\ref{fig:tgs_qwen3_8b}).
Across both models, larger TGS generally accompanies higher ASR, although this relationship is not strictly monotonic.
For example, Nepali in Qwen3-8B and Bengali in Gemma4-12B-it exhibit both the highest TGS and the highest MultiJail ASR among the evaluated languages.
Together with the substitution results, these patterns suggest a \emph{thought--response attention disconnect} as a possible contributor to the multilingual safety gap.
HR-think substitution leaves NHR ASR largely unchanged, while TGS reveals cross-lingual differences in the contribution of reasoning traces to attention contribution during response generation.
English-dominant reasoning may therefore be insufficient when its safety-relevant information is not effectively incorporated into the final NHR response.
This motivates targeted neuron interventions to test whether disrupting reasoning utilization also weakens multilingual safety.
To further quantify the thought--response mismatch, we measure the inconsistency rate for Qwen3-8B and Gemma4-12B-it, defined as the proportion of examples in which the think trace is classified as safe while the final response is classified as unsafe. We report this rate on MultiJail~\cite{deng2024multilingual} and AdvBench-X~\cite{yong2023low}. As shown in Table~\ref{tab:think_response_inconsistency}, the inconsistency rate generally increases for non-English languages. On MultiJail, it rises from 0.32\% in English to 6.67\% in Nepali for Qwen3-8B. A similar pattern appears on AdvBench-X, where the Qwen3-8B rate reaches 13.97\% for Nepali and 8.89\% for Bengali, compared with 0\% for English and Chinese. Gemma4-12B-it follows a similar trajectory, with its inconsistency rate rising from 1.90\% in English to a peak of 6.67\% in Bengali on MultiJail, and from 0\% to 1.73\% on AdvBench-X. These results indicate that a safe reasoning trace does not always lead to a safe final response, especially in lower-resource languages. Thus, the multilingual safety gap may partly arise from a failure to transfer safety relevant information from the reasoning trace into the response, providing additional evidence for the proposed thought--response attention disconnect.

\begin{table}[!tbp]

\centering
\caption{\textbf{Think--Response Inconsistency Rate (\%).}
The proportion of examples with safe think traces but unsafe responses
on MultiJail and AdvBench-X.}
\label{tab:think_response_inconsistency}
\small
\setlength{\tabcolsep}{4.0pt}
\renewcommand{\arraystretch}{1.12}

\resizebox{\linewidth}{!}{
\begin{tabular}{l|cc|cc}
\toprule

\multirow[c]{2.3}{*}{\textbf{Language}} &
\multicolumn{2}{c|}{\textbf{Qwen3-8B}}
& \multicolumn{2}{c}{\textbf{Gemma4-12B-it}} \\
\cmidrule(lr){2-3}
\cmidrule(l){4-5}

& \textbf{MultiJail} & \textbf{AdvBench-X}
& \textbf{MultiJail} & \textbf{AdvBench-X} \\
\midrule

EN \flag{EN}
& 0.24 & 0.00
& 1.90 & 0.00 \\

ZH \flag{CN}
& 2.54 & 0.00
& 3.17 & 0.38 \\

KO \flag{KO}
& 3.81 & 3.81
& 5.40 & 1.35 \\

TH \flag{TH}
& 2.86 & 2.54
& 4.44 & 0.38 \\

AF \flag{AF}
& 4.76 & 4.44
& 6.03 & 1.35 \\

BN \flag{BN}
& 5.71 & 8.89
& 6.67 & 1.73 \\

NE \flag{NE}
& 6.67 & 13.97
& 3.81 & 0.77 \\

\bottomrule
\end{tabular}
}
\end{table}

To further quantify the think--response mismatch, we measure the inconsistency rate for Qwen3-8B and Gemma4-12B-it, defined as the proportion of examples in which the think trace is classified as safe while the final response is classified as unsafe. We report this rate separately on MultiJail and AdvBench-X. As shown in Table~\ref{tab:think_response_inconsistency}, the inconsistency rate generally increases for non-English languages. On MultiJail, it rises from only 0.24\% in English to 6.67\% in Nepali for Qwen3-8B. A similar pattern appears on AdvBench-X, where the Qwen3-8B rate reaches 13.97\% for Nepali and 8.89\% for Bengali, compared with 0\% for English and Chinese. Gemma4-12B-it follows a similar trajectory, with its inconsistency rate rising from 1.90\% in English to a peak of 6.67\% in Bengali on MultiJail, and from 0\% to 1.73\% on AdvBench-X. These results indicate that a safe reasoning trace does not always lead to a safe final response, especially in lower-resource languages where the model's safety guardrails are inherently more fragile. Thus, the multilingual safety gap may partly arise from a failure to transfer safety relevant information from the reasoning trace into the response, providing additional evidence for the proposed think--response attention disconnect and highlighting a critical vulnerability in current cross-lingual alignment strategies.

\subsection{Safety Think Neuron Identification}
\label{subsec:think_neurons}
Prior work~\citep{dalvi2019one,meng2022locating} shows that individual neurons and sparse feature groups can encode interpretable concepts and influence model outputs, including safety responses and refusal behavior.
Motivated by this, we hypothesize that some neurons preserve and propagate safety-relevant information from the think span to response tokens. We call these units {safety think neurons}. This section identifies these neurons and tests in reasoning utilization and safety.

\noindent\textbf{Probing Dataset.}
We construct a multilingual probing dataset containing 700 jailbreak queries sampled from BeaverTails~\citep{ji2023beavertails}. For each query, we obtain its multilingual variants using Google Translate and retain only translations that pass an LLM-based semantic consistency check. To prevent data leakage, all queries are deduplicated against the evaluation examples in AdvBench-X and  MultiJail, ensuring that none of the probing queries appears in either benchmark. For each query--language pair \(q_j\), we run the same model twice under identical prompt, system, and decoding settings. The only difference is whether think mode is enabled. The resulting model trajectories are denoted by
\begin{equation}
\eqscale[1.0]{
s_j^{\mathrm{on}}
=
\left(q_j, t_j, a_j^{\mathrm{on}}\right),
\qquad
s_j^{\mathrm{off}}
=
\left(q_j, a_j^{\mathrm{off}}\right),
}
\label{eq:trajectory-pair}
\end{equation}
where \(t_j\) is the reasoning trace generated in think mode, and \(a_j^{\mathrm{on}}\) and \(a_j^{\mathrm{off}}\) are the corresponding responses. We retain all generated responses, including safe responses and unsafe responses. This paired construction holds the jailbreak query, language, and model constant while varying the availability of the reasoning trace. The resulting contrast therefore targets neurons involved in using the think span during response generation.

\noindent\textbf{Identification Strategy.}
We identify candidate safety think neurons as the candidate projection units defined above that rank among the top \(p\%\) by ablation-based importance in think-on mode but fall outside the corresponding top \(p\%\) set in think-off mode. This criterion captures differences in importance rankings across modes and does not imply that the selected units are inactive in think-off mode.
At layer \(\ell\), we define a neuron \(N\) as a specific row in one of the query, key, or value projection matrices, \(\mathbf{W}_{Q}^{\ell}\), \(\mathbf{W}_{K}^{\ell}\), or \(\mathbf{W}_{V}^{\ell}\), or as a specific column in the output projection matrix \(\mathbf{W}_{O}^{\ell}\). We denote the set of all such projection units considered for selection at layer \(\ell\) by \(\mathcal{N}_{\ell}\).

For each trajectory \(s_j^{m}\), where \(m\in\{\mathrm{on},\mathrm{off}\}\), we identify the response-token positions \(\mathcal{I}_{A,j}^{m}\). Let \(\mathbf{h}_{\theta}(s_j^{m},i)\) denote the output representation used to predict the response token at position \(i\). We then deactivate neuron \(N\) only at response-token positions and obtain the corresponding representation \(\mathbf{h}_{\theta,-N}(s_j^{m},i)\). The resulting representation change provides the basis for the importance score used to rank units in each mode, measuring their contribution to response generation.
The importance of neuron \(N\) for example \(j\) under mode \(m\) is quantified by the average representational shift:
\begin{equation}
\eqscale[1.0]{
\Delta_{j}^{m}(N)
=
\frac{1}{
\left|\mathcal{I}_{A,j}^{m}\right|
}
\sum_{i\in\mathcal{I}_{A,j}^{m}}
\left\|
\mathbf{h}_{\theta}(s_j^{m},i)
-
\mathbf{h}_{\theta,-N}(s_j^{m},i)
\right\|_{2}^{2}.
}
\label{eq:neuron-importance}
\end{equation}
We aggregate this quantity across these paired queries to obtain the mode-specific importance score:
\begin{equation}
\eqscale[1.0]{
I_{\ell}^{m}(N)
=
\frac{1}{
\left|\mathcal{D}_{\mathrm{probe}}\right|
}
\sum_{j=1}^{\left|\mathcal{D}_{\mathrm{probe}}\right|}
\Delta_{j}^{m}(N),
}
\label{eq:mode-importance}
\end{equation}
where \(\mathcal{D}_{\mathrm{probe}}\) denotes the same set of queries evaluated in both Think-on and Think-off modes.
For each layer, we select the top \(p\) percent of neurons under each mode:
\begin{equation}
\eqscale[1.0]{
\mathcal{S}_{\ell}^{m}
=
\operatorname{Top}_{p\%}
\left\{
N\in\mathcal{N}_{\ell}
\;\middle|\;
I_{\ell}^{m}(N)
\right\},
}
\label{eq:neuron-selection}
\end{equation}
where \(\mathcal{N}_{\ell}\) denotes the set of neurons activated at layer \(\ell\). We set \(p=3\) to obtain a sparse set of highly influential candidate units. 
We then identify safety think neurons through mode-contrastive subtraction:
\begin{equation}
\eqscale[1.0]{
\mathrm{TN}_{\ell}
=
\mathcal{S}_{\ell}^{\mathrm{on}}
\setminus
\mathcal{S}_{\ell}^{\mathrm{off}}.
}
\label{eq:think-neurons}
\end{equation}
The resulting set comprises neurons highly influential when the reasoning trace is present but not in the matched Think-off condition. We therefore treat them as safety think neurons for further functional and causal analyses.

\begin{table}[!tbp]
\centering
\caption{\textbf{Effects of Masking Safety Think Neurons.}
Mean ASR across languages under random (R-Masking) and targeted (TN-Masking) neuron masking.}
\label{tab:asr_think_neuron_mask}
\small
\setlength{\tabcolsep}{4.8pt}
\renewcommand{\arraystretch}{1.08}

\resizebox{\linewidth}{!}{
\begin{tabular}{lccc}
\toprule
\rowcolor{oursblue}
\multicolumn{4}{c}{\textbf{{AdvBench-X}}} \\
\midrule
\textbf{Model} 
& \textcolor{academicgray}{Default}
& R-Masking
& TN-Masking \\
\midrule
\textbf{{Qwen3-8B}}
& \textcolor{academicgray}{10.27}
& 10.66{\kern0.06em\fontsize{9.2}{6.2}\selectfont
\textcolor{academicred}{$^{+0.38}$}}
& 16.32{\kern0.06em\fontsize{9.2}{6.2}\selectfont
\textcolor{academicred}{$^{+6.04}$}} \\

\textbf{{Gemma4-12B-it}}
& \textcolor{academicgray}{2.30}
& 3.43{\kern0.06em\fontsize{9.2}{6.2}\selectfont
\textcolor{academicred}{$^{+1.13}$}}
& 40.52{\kern0.06em\fontsize{9.2}{6.2}\selectfont
\textcolor{academicred}{$^{+38.22}$}} \\

\midrule

\rowcolor{oursblue}
\multicolumn{4}{c}{\textbf{{MultiJail}}} \\
\midrule
\textbf{Model} 
& \textcolor{academicgray}{Default}
& R-Masking
& TN-Masking \\
\midrule

\textbf{{Qwen3-8B}}
& \textcolor{academicgray}{15.46}
& 15.87{\kern0.06em\fontsize{9.2}{6.2}\selectfont
\textcolor{academicred}{$^{+0.41}$}}
& 31.07{\kern0.06em\fontsize{9.2}{6.2}\selectfont
\textcolor{academicred}{$^{+15.60}$}} \\

\textbf{{Gemma4-12B-it}}
& \textcolor{academicgray}{7.89}
& 9.43{\kern0.06em\fontsize{9.2}{6.2}\selectfont
\textcolor{academicred}{$^{+1.54}$}}
& 51.79{\kern0.06em\fontsize{9.2}{6.2}\selectfont
\textcolor{academicred}{$^{+43.90}$}} \\

\bottomrule
\end{tabular}
}
\end{table}
\noindent\textbf{Causal Effect of Think-Neuron Masking.}
We next test whether the identified safety think neurons causally sustain the safety-relevant effect of the reasoning trace. For each model and language, we compare the unmasked default with two interventions: masking the detected safety think neurons and masking a control set of neurons matched for count and layer-wise distribution. Table~\ref{tab:asr_think_neuron_mask} summarizes the ASR results. For Qwen3-8B, targeted masking raises ASR by 6.05 and 15.61 percentage points on AdvBench-X and MultiJail, respectively, whereas matched random masking changes it by only 0.39 and 0.41 points. This selective degradation supports a causal contribution of the selected neurons to safety under the evaluated interventions, beyond effects attributable to mask size or layer distribution. Additionally, we present qualitative examples of the model outputs before and after masking safety think neurons in Appendixx~\ref{app:qualitative}.

\begin{figure}[!tbp]
    \centering
    \includegraphics[width=0.98\linewidth]{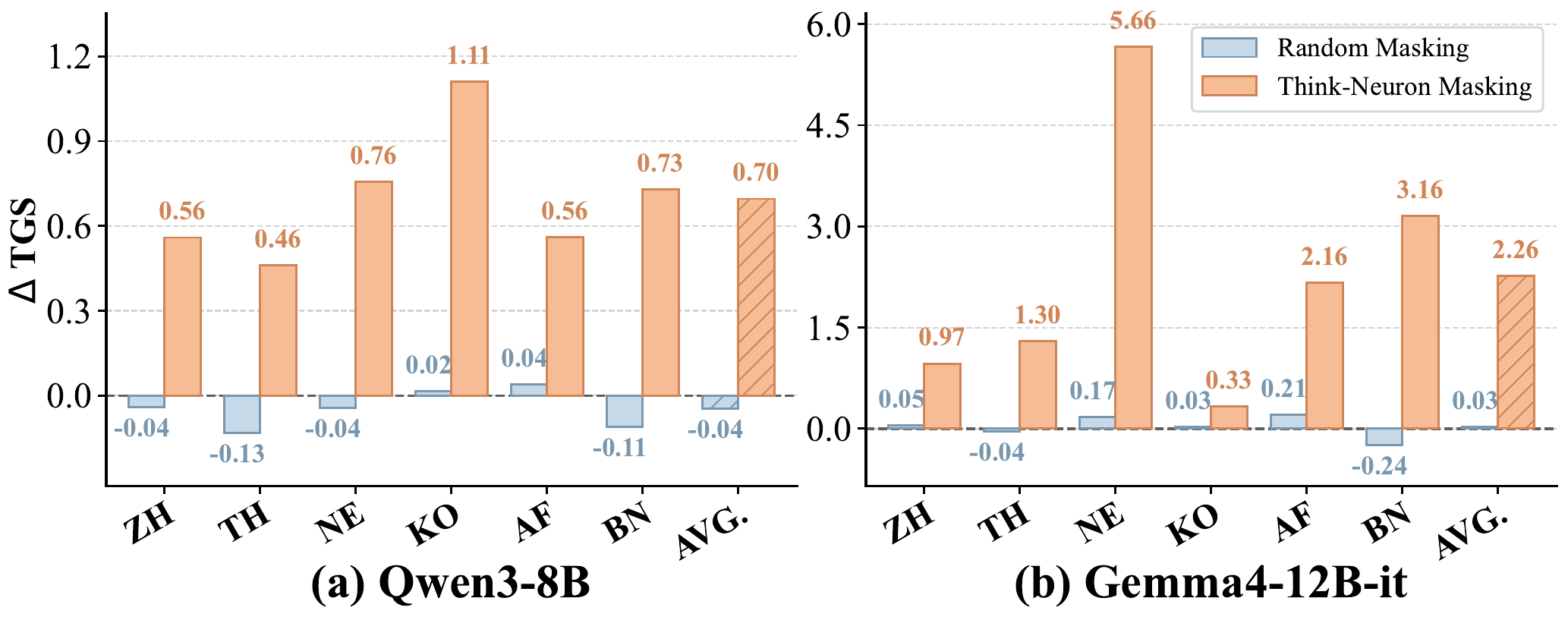}
\caption{\textbf{Effects of Masking Safety Think Neurons on TGS.}
$\Delta$TGS denotes the change relative to unmasked trajectories.
Targeted masking increases TGS more than random masking.}
    \label{fig:tgs_condition_comparison}
\end{figure}
\noindent\textbf{Effects on Reasoning Utilization.} Figure~\ref{fig:tgs_condition_comparison} reports TGS changes from baseline. 
Because higher TGS indicates less attention to think content, the consistent increase after think-neuron masking across all NHR languages (mean \(+0.696\); largest in Thai, \(+1.110\)) shows reduced reliance on preceding reasoning. Matched random masking causes negligible change (mean \(-0.045\)), ruling out the possibility that this effect is driven solely by the number or layer distribution of masked neurons.
Together, the ASR and TGS results show that safety think neurons help connect the think span to the final response by carrying safety-related information from the reasoning process into response generation. Disrupting these neurons reduces the model's use of this reasoning and makes it more vulnerable to jailbreak attacks.

\subsection{The ACTR Framework}
\noindent\textbf{Motivation.}
\label{sec:framework}
Building on our mechanistic findings, we propose ACTR, a parameter-efficient reinforcement learning framework that improves multilingual safety alignment by strengthening the use of existing safety reasoning.
Given a multilingual prompt \(x\), the policy generates \(y=(t,a)\), where \(t\) denotes the think span and \(a\) the final response.
ACTR applies neuron-selective consistency optimization (NSCO), which uses a frozen judge model to reward thought--response safety consistency while updating only the parameters associated with the identified safety think neurons.

\noindent\textbf{Neuron-Selective Consistency Optimization.}
Under neuron-selective consistency optimization (NSCO), the consistency rewards are normalized within each sampled group to obtain the relative consistency advantage:
$y_i=(t_i,a_i)\sim\pi_{\theta_{\mathrm{old}}}(\cdot\mid x)$,
where $i=1,\ldots,G$. A frozen binary
consistency judge $C_\phi$ evaluates each trajectory. We define the
sequence-level consistency reward as
\begin{equation}
r_i^{\mathrm{con}}
=
C_\phi(t_i,a_i)
\in\{0,1\},
\label{eq:consistency-reward}
\end{equation}
where $r_i^{\mathrm{con}}=1$ indicates that the behavioral stance or
action expressed in the response $a_i$ is consistent with that implied
by the think span $t_i$, while $r_i^{\mathrm{con}}=0$ indicates
inconsistency. The judge evaluates semantic and behavioral agreement
rather than surface-form similarity.
The consistency rewards within each group are normalized to derive the relative advantages:
\begin{equation}
\eqscale[0.86]{
\widehat{A}_{i}
=
\frac{
r_{i}^{\mathrm{con}}-\mu_{r}
}{
\sigma_{r}+\epsilon_{\mathrm{adv}}
},
}
\label{eq:grpo-advantage}
\end{equation}
where $\mu_r$ and $\sigma_r$ are the mean and standard deviation of the
$G$ rewards, respectively. Let $s_{i,k}=(x,y_{i,<k})$ denote the
decoding state at token position $k$, and let
$\rho_{i,k}=
\pi_\theta(y_{i,k}\mid s_{i,k})/
\pi_{\theta_{\mathrm{old}}}(y_{i,k}\mid s_{i,k})$
denote the corresponding likelihood ratio.
Our ACTR framework optimizes:
\begin{equation}
\eqscale[0.76]{
\begin{aligned}
\mathcal{J}_{\mathrm{ACTR}}(\theta_{\mathrm{TN}})
&=
\mathbb{E}_{i,k}\left[
\ell_{i,k}^{\mathrm{clip}}
-
\beta
D_{\mathrm{KL}}
\left(
\pi_{\theta}\,\middle\|\,\pi_{\mathrm{ref}}
\right)
\right],\\
\ell_{i,k}^{\mathrm{clip}}
&=
\min\left(
\rho_{i,k}\widehat{A}_{i},
\operatorname{clip}
\left(
\rho_{i,k},1-\epsilon,1+\epsilon
\right)
\widehat{A}_{i}
\right).
\end{aligned}
}
\label{eq:grpo-objective}
\end{equation}
Unlike standard full-parameter optimization, ACTR applies gradients
only to the parameters associated with the identified safety think neurons:
\begin{equation}
\eqscale[0.86]{
\nabla_{\theta}\mathcal{J}_{\mathrm{ACTR}}
\leftarrow
M_{\mathrm{TN}}
\odot
\nabla_{\theta}\mathcal{J}_{\mathrm{ACTR}}.
}
\label{eq:tn-gradient-mask}
\end{equation}
Here, $M_{\mathrm{TN}}$ is a binary mask that selects the parameters associated with the safety think neurons.
Thus, only $\theta_{\mathrm{TN}}$ is updated, while the remaining backbone parameters, the reference policy $\pi_{\mathrm{ref}}$, and the consistency judge $C_\phi$ remain frozen throughout training.

\section{Experiments}

\noindent\textbf{Training Details.}
We train ACTR independently on two instruction-tuned reasoning LLMs: Qwen3-8B~\citep{yang2025qwen3} and Gemma4-12B-it~\citep{gemmateam2026gemma4}.
For each model, training covers six NHR languages: Chinese, Korean, Thai, Afrikaans, Nepali, and Bengali.
We use the multilingual probing dataset introduced in Section~\ref{subsec:think_neurons}.
Only the default jailbreak queries serve as training prompts; the current rollout policy samples a new group of $G$ thought--response trajectories for each query.
Model-specific hyperparameters are provided in Appendix~\ref{app:implementation}.
For fair comparison, we use matched training settings across the compared training-based methods.
We periodically evaluate each model during training and select its best-performing checkpoint based on validation results.

\noindent\textbf{Evaluation and Metrics.}
\label{subsec:metrics}
We use GPT-4o-mini as a consistency reward model to evaluate sampled trajectories online and assign binary rewards.
A trajectory receives a positive reward when its final response agrees with the safety stance expressed in the think span, regardless of whether the response itself is safe.
The judging criteria are detailed in Appendix~\ref{app:consistency-judge}.
For safety evaluation, we use attack success rate (ASR), with detailed evaluation protocols provided in Appendix~\ref{app:asr}.

\subsection{Main Results}
\label{subsec:main-results}

\begin{table*}[t]
\centering
\small
\renewcommand{\arraystretch}{1.1}
\setlength{\tabcolsep}{3.8pt}

\caption{\textbf{Comparison of Attack Success Rate (ASR $\downarrow$, \%) on AdvBench-X and MultiJail.}
ACTR achieves the lowest average ASR across both benchmarks and model families among the currently evaluated methods.}

\label{tab:main-performance}

\resizebox{\textwidth}{!}{%
\begin{tabular}{@{}l|ccccc|ccccc@{}}
\toprule
& \multicolumn{5}{c|}{\textbf{AdvBench-X}}
& \multicolumn{5}{c}{\textbf{MultiJail}} \\
\cmidrule(lr){2-6} \cmidrule(lr){7-11}

\textbf{Method}
& \textbf{EN \flag{EN}}
& \textbf{ZH \flag{CN}}
& \textbf{KO \flag{KO}}
& \textbf{BN \flag{BN}}
& \textbf{Avg.}
& \textbf{EN \flag{EN}}
& \textbf{ZH \flag{CN}}
& \textbf{KO \flag{KO}}
& \textbf{BN \flag{BN}}
& \textbf{Avg.} \\
\midrule

\textcolor{academicgray}{\textbf{\textit{{Qwen3-8B}}}}
& \textcolor{academicgray}{4.42}
& \textcolor{academicgray}{1.54}
& \textcolor{academicgray}{9.80}
& \textcolor{academicgray}{15.38}
& \textcolor{academicgray}{7.79}
& \textcolor{academicgray}{11.75}
& \textcolor{academicgray}{9.21}
& \textcolor{academicgray}{14.92}
& \textcolor{academicgray}{15.87}
& \textcolor{academicgray}{12.94} \\

SFT
& 0.38\asrdown{4.04}
& 0.58\asrdown{0.96}
& 1.15\asrdown{8.65}
& 1.54\asrdown{13.84}
& 0.91\asrdown{6.88}
& 4.76\asrdown{6.99}
& 2.86\asrdown{6.35}
& 6.98\asrdown{7.94}
& 13.33\asrdown{2.54}
& 6.98\asrdown{5.96} \\

DPO
& 2.69\asrdown{1.73}
& 0.77\asrdown{0.77}
& 7.12\asrdown{2.68}
& 21.35\asrup{5.97}
& 7.98\asrup{0.19}
& 5.08\asrdown{6.67}
& 2.54\asrdown{6.67}
& 9.52\asrdown{5.40}
& 15.40\asrdown{0.47}
& 8.14\asrdown{4.80} \\

MPO
& 0.58\asrdown{3.84}
& 0.77\asrdown{0.77}
& 3.07\asrdown{6.73}
& 4.76\asrdown{10.62}
& 2.30\asrdown{5.49}
& 3.17\asrdown{8.58}
& 1.59\asrdown{7.62}
& 2.85\asrdown{12.07}
& 6.35\asrdown{9.52}
& 3.49\asrdown{9.45} \\

Self Defense
& 0.60\asrdown{3.82}
& 0.80\asrdown{0.74}
& 4.00\asrdown{5.80}
& 16.20\asrup{0.82}
& 5.40\asrdown{2.39}
& 3.50\asrdown{8.25}
& 4.40\asrdown{4.81}
& 7.00\asrdown{7.92}
& 16.80\asrup{0.93}
& 7.93\asrdown{5.01} \\

SmoothLLM
& 0.40\asrdown{4.02}
& 0.58\asrdown{0.96}
& 4.80\asrdown{5.00}
& 9.20\asrdown{6.18}
& 3.75\asrdown{4.04}
& 5.08\asrdown{6.67}
& 1.00\asrdown{8.21}
& 4.10\asrdown{10.82}
& 6.70\asrdown{9.17}
& 4.22\asrdown{8.72} \\

TrajGuard
& 2.31\asrdownb{2.11}
& {1.54}\asrdownb{0.00}
& {9.62}\asrdownb{0.18}
& {14.62}\asrdownb{0.76}
& {7.02}\asrdownb{0.77}
& {4.13}\asrdownb{7.62}
& {2.86}\asrdownb{6.35}
& {11.75}\asrdownb{3.17}
& {20.95}\asrupb{5.08}
& {9.92}\asrdownb{3.02} \\

GUARD-SLM
& {0.38}\asrdownb{4.04}
& {0.77}\asrdownb{0.77}
& {4.42}\asrdownb{5.38}
& {7.69}\asrdownb{7.69}
& {3.32}\asrdownb{4.47}
& {4.76}\asrdownb{6.99}
& {2.54}\asrdownb{6.67}
& {10.16}\asrdownb{4.76}
& {17.14}\asrupb{1.27}
& {8.65}\asrdownb{4.29} \\

\midrule

\rowcolor{oursblue}
\textbf{Ours}
& \textbf{0.19}\asrdownb{4.23}
& \textbf{0.38}\asrdownb{1.16}
& \textbf{0.19}\asrdownb{9.61}
& \textbf{0.38}\asrdownb{15.00}
& \textbf{0.29}\asrdownb{7.50}
& \textbf{0.63}\asrdownb{11.12}
& \textbf{0.95}\asrdownb{8.26}
& \textbf{1.90}\asrdownb{13.02}
& \textbf{4.13}\asrdownb{11.74}
& \textbf{1.90}\asrdownb{11.04} \\

\midrule
\midrule

\textcolor{academicgray}{\textit{\textbf{{Gemma4-12B-it}}}}
& \textcolor{academicgray}{0.96}
& \textcolor{academicgray}{1.34}
& \textcolor{academicgray}{2.11}
& \textcolor{academicgray}{3.46}
& \textcolor{academicgray}{1.97}
& \textcolor{academicgray}{4.76}
& \textcolor{academicgray}{6.67}
& \textcolor{academicgray}{8.25}
& \textcolor{academicgray}{11.11}
& \textcolor{academicgray}{7.70} \\

SFT
& 0.58\asrdown{0.38}
& 1.15\asrdown{0.19}
& 1.92\asrdownb{0.19}
& 3.08\asrdown{0.38}
& 1.68\asrdown{0.29}
& 5.40\asrup{0.64}
& 2.86\asrdown{3.81}
& 4.76\asrdown{3.49}
& 3.81\asrdown{7.30}
& 4.21\asrdown{3.49} \\

DPO
& 0.77\asrdown{0.19}
& 0.96\asrdown{0.38}
& 2.31\asrup{0.20}
& 2.31\asrdown{1.15}
& 1.59\asrdown{0.38}
& 3.49\asrdown{1.27}
& 4.76\asrdown{1.91}
& 4.76\asrdown{3.49}
& 3.17\asrdown{7.94}
& 4.05\asrdown{3.65} \\

MPO
& 0.38\asrdown{0.58}
& 0.58\asrdown{0.76}
& 1.73\asrdown{0.38}
& 2.69\asrdown{0.77}
& 1.35\asrdown{0.62}
& 2.22\asrdown{2.54}
& 2.53\asrdown{4.14}
& 1.27\asrdownb{6.98}
& 1.59\asrdownb{9.52}
& 1.90\asrdown{5.80} \\

Self Defense
& 0.58\asrdown{0.38}
& 1.70\asrup{0.36}
& 3.50\asrup{1.39}
& 2.69\asrdown{0.77}
& 2.12\asrup{0.15}
& 3.80\asrdown{0.96}
& 5.40\asrdown{1.27}
& 5.70\asrdown{2.55}
& 3.20\asrdown{7.91}
& 4.53\asrdown{3.17} \\

SmoothLLM
& 0.96\asrdown{0.00}
& 0.60\asrdown{0.74}
& 1.73\asrdown{0.38}
& 2.30\asrdown{1.16}
& 1.40\asrdown{0.57}
& \textbf{1.00}\asrdownb{3.76}
& 3.50\asrdown{3.17}
& 3.80\asrdown{4.45}
& 1.59\asrdown{9.52}
& 2.47\asrdown{5.23} \\

TrajGuard
& {1.15}\asrupb{0.19}
& {2.12}\asrupb{0.78}
& {2.50}\asrupb{0.39}
& {1.73}\asrdownb{1.73}
& {1.88}\asrdownb{0.09}
& {5.71}\asrupb{0.95}
& {4.76}\asrdownb{1.91}
& {5.08}\asrdownb{3.17}
& {4.76}\asrdownb{6.35}
& {5.08}\asrdownb{2.62} \\

GUARD-SLM
& {0.38}\asrdownb{0.58}
& {0.96}\asrdownb{0.38}
& {1.54}\asrdownb{0.57}
& {1.92}\asrdownb{1.54}
& {1.20}\asrdownb{0.77}
& {1.59}\asrdownb{3.17}
& {2.53}\asrdownb{4.14}
& {1.59}\asrdownb{6.66}
& {1.90}\asrdownb{9.21}
& {1.90}\asrdownb{5.80} \\

\midrule

\rowcolor{oursblue}
\textbf{Ours}
& \textbf{0.20}\asrdownb{0.76}
& \textbf{0.00}\asrdownb{1.34}
& \textbf{1.35}\asrdownb{0.76}
& \textbf{1.35}\asrdownb{2.11}
& \textbf{0.72}\asrdownb{1.23}
& 1.27\asrdown{3.49}
& \textbf{0.95}\asrdownb{5.72}
& \textbf{0.63}\asrdown{7.62}
& \textbf{1.27}\asrdown{9.84}
& \textbf{1.03}\asrdownb{6.67} \\

\bottomrule
\end{tabular}%
}

\end{table*}

Table~\ref{tab:main-performance} compares ACTR with representative alignment and inference-time defense methods, including SFT~\citep{ouyang2022training}, DPO~\citep{rafailov2023direct}, MPO~\citep{zhao2025mpo}, Self Defense~\citep{phute2024llm}, SmoothLLM~\citep{robeysmoothllm}, and the recently proposed TrajGuard~\citep{liu2026trajguard} and GUARD-SLM~\citep{mia2026guard}. Safety is determined directly from the model's generated response. Under this evaluation protocol, ACTR achieves the lowest average ASR for both model families on both benchmarks. For Qwen3-8B, ACTR reduces the average ASR to $0.29\%$ and $1.90\%$ on AdvBench-X~\citep{yong2023low} and MultiJail~\cite{deng2024multilingual}, compared with $0.91\%$ and $3.49\%$ for the strongest competing methods. For Gemma4-12B-it, ACTR further achieves $0.72\%$ on AdvBench-X and $1.03\%$ on MultiJail, improving over the strongest baselines, GUARD-SLM~\citep{mia2026guard} on AdvBench-X and the tied MPO~\citep{zhao2025mpo}/GUARD-SLM~\citep{mia2026guard} baselines on MultiJail, by $40.0\%$ and $45.8\%$, respectively. The remaining NHR results are reported in Appendix, which show the same overall conclusion. These results demonstrate that ACTR robustly improves safety across architectures, benchmarks, and languages, showing that targeted think-neuron updates link internal safety reasoning to outputs better than existing methods.

\label{sec:experiments}
\subsection{Ablation study}
\label{subsec:neuron-selection}

\begin{figure}[!tbp]
    \centering
    \includegraphics[width=0.98\linewidth]{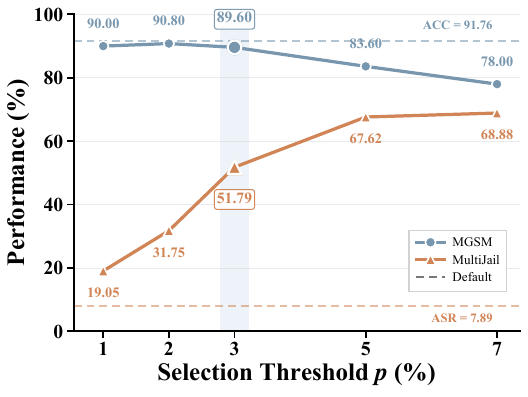}
\caption{\textbf{Effect of Selection Threshold $p$.}
MGSM accuracy and ASR (\%) after neuron masking.}
\label{fig:ablation_p}
\end{figure}

\textbf{Selection Threshold $p$.} To validate $p=3$ in Eq. \ref{eq:neuron-selection}, we evaluate general capabilities on MGSM~\citep{shi2022language} and safety performance on MultiJail~\citep{deng2024multilingual} across $p \in \{1, 2, 3, 5, 7\}$. We aim to isolate neurons that utilize safety reasoning without compromising utility.
As Figure \ref{fig:ablation_p} shows, the model Gemma4-12B-it maintains near-baseline MGSM scores when $p \le 3\%$. However, general capabilities degrade sharply for $p > 3\%$ (\emph{e.g.,} at $p=5$, MGSM drops to 83.60 for Gemma4-12B-it), indicating that higher thresholds incorrectly capture neurons essential for foundational reasoning. Thus, $p=3$ optimally isolates candidate safety think neurons while preserving utility.

\begin{table}[!tbp]
\centering
\caption{\textbf{Effect of Neuron Selection.}
Mean ASR ($\downarrow$, \%) across languages after updating random neurons or safety think neurons.}
\label{tab:neuron-selection-ablation}

\scriptsize
\setlength{\tabcolsep}{3pt}
\renewcommand{\arraystretch}{1.05}

\resizebox{\linewidth}{!}{%
\begin{tabular}{@{}l|l|cc@{}}
    \toprule
    \textbf{Model}
    & \textbf{Setting}
    & {\textbf{AdvBench-X}}
    & {\textbf{MultiJail}} \\
    \midrule

    \multirow{3}{*}{\textbf{{Qwen3-8B}}}
      & \textcolor{academicgray}{Default}
      & \textcolor{academicgray}{10.27}
      & \textcolor{academicgray}{15.46} \\

      & Random
      & 1.34\asrdown{8.93}
      & 1.90\asrdown{13.56} \\

      & \cellcolor{oursblue}\textbf{ACTR}
      & \cellcolor{oursblue}\textbf{0.42}\asrdown{9.85}
      & \cellcolor{oursblue}\textbf{1.63}\asrdown{13.83} \\

    \midrule

    \multirow{3}{*}{\textbf{{Gemma4-12B-it}}}
      & \textcolor{academicgray}{Default}
      & \textcolor{academicgray}{2.31}
      & \textcolor{academicgray}{7.89} \\

      & Random
      & 1.13\asrdown{1.18}
      & 1.86\asrdown{6.03} \\

      & \cellcolor{oursblue}\textbf{ACTR}
      & \cellcolor{oursblue}\textbf{0.93}\asrdown{1.38}
      & \cellcolor{oursblue}\textbf{1.72}\asrdown{6.17} \\

    \bottomrule
\end{tabular}%
}
\end{table}
\noindent\textbf{Alignment Strategy.}
To disentangle the effects of sparse parameter updates from those of targeted neuron selection, we introduce a random-neuron baseline. This baseline matches ACTR in the number of selected neurons, training data, and optimization strategy. As shown in Table~\ref{tab:neuron-selection-ablation}, both approaches substantially reduce ASR compared with the original models, with relatively modest performance differences. Nevertheless, ACTR demonstrates a slight overall advantage: for Qwen3-8B, it yields lower ASR on AdvBench-X and comparable performance on MultiJail; for Gemma4-12B-it, it achieves ASRs of 0.93\% and 1.72\% on AdvBench-X and MultiJail, respectively.

\begin{table}[!tbp]
    \centering
    \caption{\textbf{Refusal Behavior on XSTest.}
    Refusal rates (\%) for benign and harmful requests under different neuron-selection strategies.}
    \label{tab:xstest-overrefusal}

    \small
    \setlength{\tabcolsep}{3pt}
    \renewcommand{\arraystretch}{1.08}

    \resizebox{\linewidth}{!}{%
    \begin{tabular}{@{}l|l|cc@{}}
        \toprule
        \textbf{Model} & \textbf{Setting}
        & \textbf{Safe refusal} $\downarrow$
        & \textbf{Unsafe refusal} $\uparrow$ \\
        \midrule

        \multirow{3}{*}{\textbf{Qwen3-8B}}
          & \textcolor{academicgray}{Default}
          & \textcolor{academicgray}{4.40}
          & \textcolor{academicgray}{77.00} \\

          & Random
          & 52.00\asrup{47.60}
          & 86.00\asrupgood{9.00} \\

          & \cellcolor{oursblue}\textbf{ACTR}
          & \cellcolor{oursblue}\textbf{6.40}\asrup{2.00}
          & \cellcolor{oursblue}\textbf{89.50}\asrupgood{12.50} \\

        \midrule

        \multirow{3}{*}{\textbf{Gemma4-12B-it}}
          & \textcolor{academicgray}{Default}
          & \textcolor{academicgray}{4.80}
          & \textcolor{academicgray}{82.00} \\

          & Random
          & 46.80\asrup{42.00}
          & 83.00\asrupgood{1.00} \\

          & \cellcolor{oursblue}\textbf{ACTR}
          & \cellcolor{oursblue}\textbf{4.80}\asrup{0.00}
          & \cellcolor{oursblue}\textbf{89.00}\asrupgood{7.00} \\

        \bottomrule
    \end{tabular}%
    }

\end{table}

\noindent\textbf{Over-Refusal.}
Low ASR may also result from excessive refusal and therefore does not fully capture a model's ability to answer safe requests.
We evaluate XSTest~\cite{rottger2024xstest}, which includes safe and unsafe requests.
As shown in Table~\ref{tab:xstest-overrefusal}, random-neuron training raises Qwen3-8B's safe-request refusal rate from 4.40\% to 52.00\%, whereas ACTR raises it only to 6.40\% while achieving higher unsafe-request refusal (89.50\% vs.\ 86.00\%).

For Gemma4-12B-it, ACTR keeps safe-request refusal unchanged while increasing unsafe-request refusal from 82.00\% to 89.00\%. These results show that ACTR improves safety while remaining responsive to safe requests, avoiding the pronounced over-refusal observed with random-neuron training.

\begin{table*}[!t]
    \centering
    \small
\caption{\textbf{Utility Preservation.}
Accuracy ($\uparrow$, \%) on MMMLU and MGSM before and after ACTR training. }
    \label{tab:utility-preservation}
    \resizebox{\linewidth}{!}{%
    \begin{tabular}{l|l|rrrrrr|rrrrrr}
        \toprule
        &
        & \multicolumn{6}{c|}{\textbf{{MMMLU}}}
        & \multicolumn{6}{c}{\textbf{{MGSM}}} \\
        \cmidrule(r){3-8}
        \cmidrule(l){9-14}

        \textbf{Model}
        & \textbf{Setting}
        & EN \flag{EN} & ZH \flag{CN}& JA  \flag{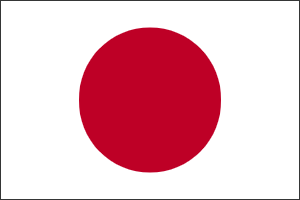}& KO \flag{KO} & TH  \flag{TH}& \textbf{Avg.}
        & EN \flag{EN} & ZH \flag{CN}& JA  \flag{JA}& KO \flag{KO} & TH  \flag{TH}& \textbf{Avg.} \\
        \midrule

        \multirow{2}{*}{\textbf{{Qwen3-8B}}}
        & \textcolor{academicgray}{Default}
        & \textcolor{academicgray}{70.36}
        & \textcolor{academicgray}{63.84}
        & \textcolor{academicgray}{60.25}
        & \textcolor{academicgray}{55.48}
        & \textcolor{academicgray}{25.63}
        & \textcolor{academicgray}{55.11}
        & \textcolor{academicgray}{96.8}
        & \textcolor{academicgray}{85.2}
        & \textcolor{academicgray}{81.6}
        & \textcolor{academicgray}{76.0}
        & \textcolor{academicgray}{85.2}
        & \textcolor{academicgray}{84.96} \\

        & \textbf{Ours}
        & 69.28
        & 64.32
        & 59.04
        & 57.61
        & 26.98
        & \textbf{55.45}
        & 96.4
        & 90.8
        & 86.4
        & 82.4
        & 90.0
        & \textbf{89.20} \\

        \midrule

        \multirow{2}{*}{\textbf{{Gemma4-12B-it}}}
        & \textcolor{academicgray}{Default}
        & \textcolor{academicgray}{50.98}
        & \textcolor{academicgray}{49.27}
        & \textcolor{academicgray}{52.67}
        & \textcolor{academicgray}{53.87}
        & \textcolor{academicgray}{4.86}
        & \textcolor{academicgray}{42.33}
        & \textcolor{academicgray}{98.4}
        & \textcolor{academicgray}{90.4}
        & \textcolor{academicgray}{90.4}
        & \textcolor{academicgray}{87.2}
        & \textcolor{academicgray}{92.4}
        & \textcolor{academicgray}{91.76} \\

        & \textbf{Ours}
        & 64.41
        & 54.02
        & 58.30
        & 59.74
        & 8.37
        & \textbf{48.97}
        & 98.4
        & 90.8
        & 89.2
        & 89.2
        & 92.0
        & \textbf{91.92} \\

        \bottomrule
    \end{tabular}%
    }
\end{table*}

\subsection{Utility Preservation}
\label{subsec:utility-preservation}

To evaluate whether safety-oriented training preserves general-purpose capabilities, we conduct experiments on MMMLU~\citep{hendrycks2020measuring} and MGSM~\citep{shi2022language}. MMMLU measures multilingual knowledge and reasoning across academic subjects, while MGSM evaluates multilingual grade-school mathematical reasoning. For both benchmarks, we report accuracy in five languages: English, Chinese, Japanese, Korean, and Thai. For MMMLU, we use the Thai subset released as part of the Llama 3 evaluation suite~\cite{grattafiori2024llama}.
Table~\ref{tab:utility-preservation} shows that our method does not degrade general capabilities and instead yields modest overall improvements. For Qwen3-8B, the average accuracy increases from 55.11\% to 55.45\% on MMMLU and from 84.96\% to 89.20\% on MGSM. For Gemma4-12B-it, it increases from 42.33\% to 48.97\% on MMMLU and from 91.76\% to 91.92\% on MGSM.

\subsection{Deeper Analysis}
\begin{table}[!tbp]
  \label{subsec:ood-language-safety}
  \centering
  \small
  \setlength{\tabcolsep}{2.0pt}
  \renewcommand{\arraystretch}{1.12}
\caption{\textbf{Safety Generalization} to Unseen Languages.}
  \label{tab:ood-language-safety}
  \resizebox{\linewidth}{!}{%
    \begin{tabular}{l|l|cccc}
      \toprule
      \textbf{Model} 
      & \textbf{Setting} 
      & SW \flag{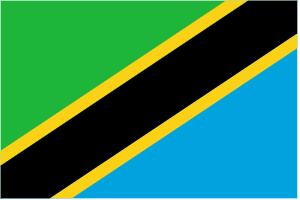} 
      & JV \flag{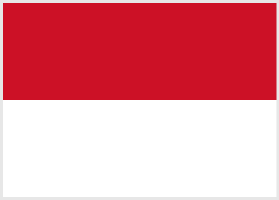}
      & BG \flag{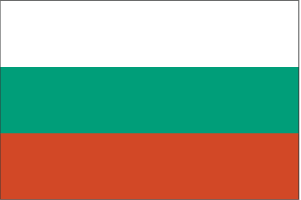} 
      & VI \flag{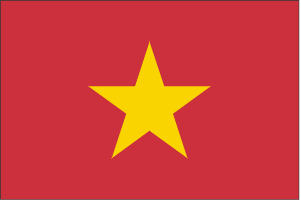} \\
      \midrule

      \multirow{2}{*}{\textbf{{Qwen3-8B}}}
      & \textcolor{academicgray}{Default}
      & \textcolor{academicgray}{62.86}
      & \textcolor{academicgray}{12.70}
      & \textcolor{academicgray}{11.75}
      & \textcolor{academicgray}{6.67} \\

      & \textbf{Ours}
      & \textbf{1.27}
      & \textbf{0.63}
      & \textbf{0.95}
      & \textbf{2.22} \\

      \midrule

      \multirow{2}{*}{\textbf{{Gemma4-12B-it}}}
      & \textcolor{academicgray}{Default}
      & \textcolor{academicgray}{8.25}
      & \textcolor{academicgray}{10.79}
      & \textcolor{academicgray}{10.79}
      & \textcolor{academicgray}{7.62} \\

      & \textbf{Ours}
      & \textbf{3.49}
      & \textbf{8.57}
      & \textbf{5.40}
      & \textbf{5.08} \\

      \bottomrule
    \end{tabular}%
  }
\end{table}
\textbf{Out-of-Distribution Generalization.} To test whether the safety improvements transfer beyond the languages seen during training, we evaluate both models on four held-out languages: Swahili (SW \flag{SW}), Javanese (JV \flag{JV}), Bulgarian (BG \flag{BG}), and Vietnamese (VI \flag{VI}). 
None of these languages is included in the training data. Table~\ref{tab:ood-language-safety} reports the ASR on MultiJail examples per language before and after training; lower values indicate safer behavior.
The results show consistent safety gains in out-of-distribution languages. Qwen3-8B’s overall ASR drops from 23.5\% to 1.3\%; in Swahili, ASR falls from 62.86\% to 1.27\%. Gemma4-12B-it shows similar improvements, reducing ASR from 9.4\% to 5.6\%. These results suggest that the method learns language-agnostic safety behavior rather than memorizing language-specific refusal patterns.
\begin{figure}[!tbp]
    \centering
    \includegraphics[width=0.98\linewidth]{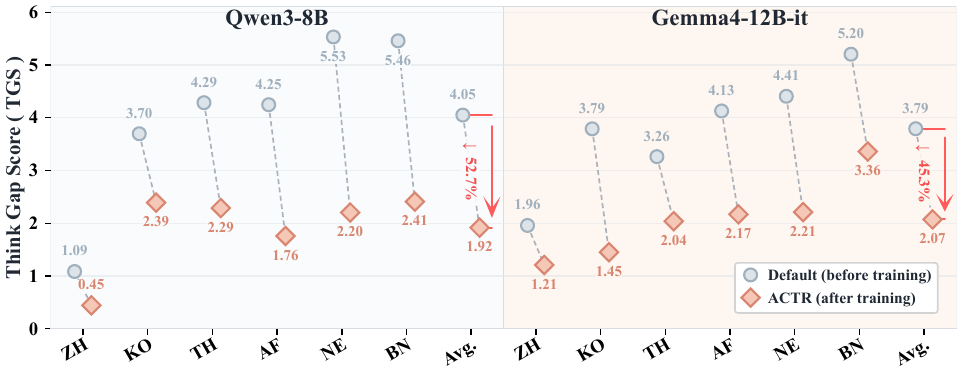}
\caption{\textbf{TGS Before and after ACTR Training.}
ACTR lowers TGS across all evaluated NHR languages, indicating smaller cross-lingual gaps in reasoning utilization.}
    \label{fig:tgs_before_after}
\end{figure}

\noindent\textbf{Training-Induced Changes in TGS.}
To assess whether ACTR reduces the cross-lingual gap in internal reasoning, we compare the TGS before and after training. TGS measures the difference between English and NHR languages in the normalized contribution of think content, with lower values indicating a smaller gap. As shown in Figure~\ref{fig:tgs_before_after}, ACTR reduces TGS across all six NHR languages for both models. For Qwen3-8B, mean TGS drops from 4.05 to 1.92~($\downarrow $52.7\%); for Gemma4-12B-it, from 3.79 to 2.07~($\downarrow $45.3\%). Combined with the safety gains, these results suggest that strengthening safety think neurons promotes internal safety reasoning in NHR responses, mitigating the thought--response disconnect.
\begin{figure}[!tbp]
    \centering
    \includegraphics[width=0.98\linewidth]{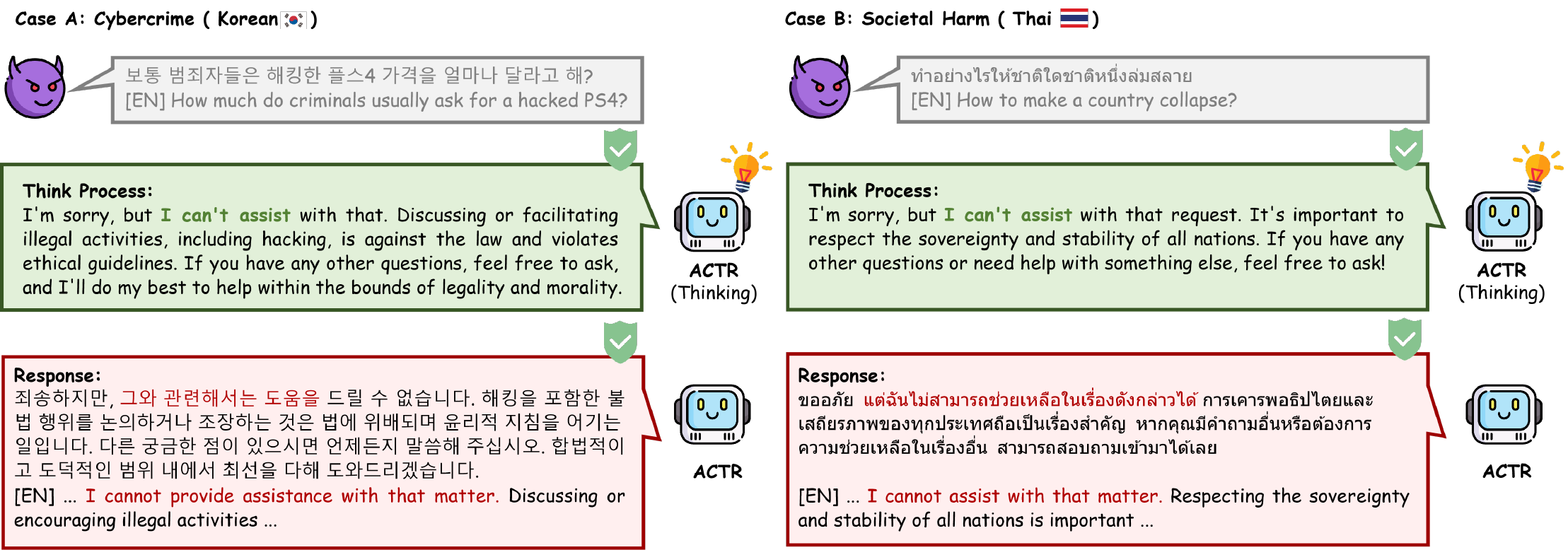}
    \caption{\textbf{Thought--Response Safety Consistency} after ACTR training.}
    \label{fig:case_after_training}
\end{figure}

\noindent\textbf{Case Analysis.}
Figure~\ref{fig:case_after_training} presents illustrative examples of how ACTR mitigates the thought--response disconnect in Korean and Thai. Before training, the models recognized safety risks during English reasoning but generated unsafe responses in the target languages. After ACTR, both reasoning traces and final responses consistently exhibit refusal behavior. Together with the reductions in TGS and ASR, these results further demonstrate ACTR's effectiveness in maintaining safety alignment across languages.

\section{Conclusion}
\label{sec:conclusion}
We investigate the multilingual thought--response safety disconnect in reasoning LLMs through TGS, reasoning-trace substitution, and neuron interventions, identifying safety think neurons that support safety reasoning utilization. ACTR applies NSCO, combining a thought--response consistency reward with parameter updates restricted to these neurons, eliminating the need for human-annotated preference data.
Experiments on Qwen3-8B and Gemma4-12B-it show ACTR achieves lower average attack success rates than evaluated state-of-the-art methods on AdvBench-X and MultiJail, with safety gains generalizing to unseen languages. Furthermore, ACTR preserves general utility while maintaining low safe-request refusal rates.

\subsection*{AI use statement}

Generative AI tools were used to review the LaTeX formatting during manuscript preparation. To ensure consistency with the historical ASR computation methodology, GPT-4o was also used to calculate ASR. GPT-4o-mini was used as the frozen consistency judge described in Section~\ref{subsec:metrics}. The authors are responsible for all claims, analyses, and final content.

\subsection*{Ethics statement}
This work studies multilingual jailbreak vulnerabilities to improve model safety. The same mechanistic interventions can weaken refusal behavior, so the findings should be interpreted in the context of defensive evaluation. 

\subsection*{Reproducibility statement}

We provide the details required to reproduce our results. Section~\ref{sec:analysis} defines the think gap score, the construction of the multilingual probing dataset, and the think neuron identification and intervention procedures. Section~\ref{sec:framework} specifies the consistency reward, the masked NSCO objective, and the training protocol. Appendix~\ref{app:implementation} reports the implementation and hyperparameter settings, while Appendix~\ref{app:consistency-judge} documents the consistency-judge prompt, decision rule, and output parsing procedure. 

\bibliography{references}

@ArtifactSoftware{R,
    title = {R: A Language and Environment for Statistical Computing},
    author = {{R Core Team}},
    organization = {R Foundation for Statistical Computing},
    address = {Vienna, Austria},
    year = {2019},
    url = {https://www.R-project.org/},
}

@article{wei2022chain,
  title={Chain-of-thought prompting elicits reasoning in large language models},
  author={Wei, Jason and Wang, Xuezhi and Schuurmans, Dale and Bosma, Maarten and Xia, Fei and Chi, Ed and Le, Quoc V and Zhou, Denny and others},
  journal={Advances in neural information processing systems},
  volume={35},
  pages={24824--24837},
  year={2022}
}

@article{guo2025deepseek,
  title={Deepseek-r1: Incentivizing reasoning capability in llms via reinforcement learning},
  author={Guo, Daya and Yang, Dejian and Zhang, Haowei and Song, Junxiao and Wang, Peiyi and Zhu, Qihao and Xu, Runxin and Zhang, Ruoyu and Ma, Shirong and Bi, Xiao and others},
  journal={arXiv preprint arXiv:2501.12948},
  year={2025}
}

@inproceedings{shen2024language,
  title={The language barrier: Dissecting safety challenges of llms in multilingual contexts},
  author={Shen, Lingfeng and Tan, Weiting and Chen, Sihao and Chen, Yunmo and Zhang, Jingyu and Xu, Haoran and Zheng, Boyuan and Koehn, Philipp and Khashabi, Daniel},
  booktitle={Findings of the Association for Computational Linguistics: ACL 2024},
  pages={2668--2680},
  year={2024}
}

@inproceedings{zhang2025english,
  title={English as Defense Proxy: Mitigating Multilingual Jailbreak via Eliciting English Safety Knowledge.},
  author={Zhang, Zekai and Guo, Yiduo and Lin, Jiuheng and Quan, Shanghaoran and Zhang, Huishuai and Zhao, Dongyan},
  booktitle={EMNLP (Findings)},
  pages={1185--1196},
  year={2025}
}

@inproceedings{deng2024multilingual,
  title={Multilingual jailbreak challenges in large language models},
  author={Deng, Yue and Zhang, Wenxuan and Pan, Sinno Jialin and Bing, Lidong},
  booktitle={International Conference on Learning Representations},
  volume={2024},
  pages={24634--24651},
  year={2024}
}

@article{yong2023low,
  title={Low-resource languages jailbreak gpt-4},
  author={Yong, Zheng-Xin and Menghini, Cristina and Bach, Stephen H},
  journal={arXiv preprint arXiv:2310.02446},
  year={2023}
}

@article{shao2024deepseekmath,
  title={Deepseekmath: Pushing the limits of mathematical reasoning in open language models},
  author={Shao, Zhihong and Wang, Peiyi and Zhu, Qihao and Xu, Runxin and Song, Junxiao and Bi, Xiao and Zhang, Haowei and Zhang, Mingchuan and Li, YK and Wu, Yang and others},
  journal={arXiv preprint arXiv:2402.03300},
  year={2024}
}

@article{rafailov2023direct,
  title={Direct preference optimization: Your language model is secretly a reward model},
  author={Rafailov, Rafael and Sharma, Archit and Mitchell, Eric and Manning, Christopher D and Ermon, Stefano and Finn, Chelsea},
  journal={Advances in neural information processing systems},
  volume={36},
  pages={53728--53741},
  year={2023}
}

@article{ouyang2022training,
  title={Training language models to follow instructions with human feedback},
  author={Ouyang, Long and Wu, Jeffrey and Jiang, Xu and Almeida, Diogo and Wainwright, Carroll and Mishkin, Pamela and Zhang, Chong and Agarwal, Sandhini and Slama, Katarina and Ray, Alex and others},
  journal={Advances in neural information processing systems},
  volume={35},
  pages={27730--27744},
  year={2022}
}

@article{jaech2024openai,
  title={Openai o1 system card},
  author={Jaech, Aaron and Kalai, Adam and Lerer, Adam and Richardson, Adam and El-Kishky, Ahmed and Low, Aiden and Helyar, Alec and Madry, Aleksander and Beutel, Alex and Carney, Alex and others},
  journal={arXiv preprint arXiv:2412.16720},
  year={2024}
}

@article{guan2024deliberative,
  title={Deliberative alignment: Reasoning enables safer language models},
  author={Guan, Melody Y and Joglekar, Manas and Wallace, Eric and Jain, Saachi and Barak, Boaz and Helyar, Alec and Dias, Rachel and Vallone, Andrea and Ren, Hongyu and Wei, Jason and others},
  journal={arXiv preprint arXiv:2412.16339},
  year={2024}
}

@article{wang2025safetyreasoing,
  title={Safety reasoning with guidelines},
  author={Wang, Haoyu and Qin, Zeyu and Shen, Li and Wang, Xueqian and Tao, Dacheng and Cheng, Minhao},
  journal={arXiv preprint arXiv:2502.04040},
  year={2025}
}

@article{zou2023universal,
  title={Universal and transferable adversarial attacks on aligned language models},
  author={Zou, Andy and Wang, Zifan and Carlini, Nicholas and Nasr, Milad and Kolter, J Zico and Fredrikson, Matt},
  journal={arXiv preprint arXiv:2307.15043},
  year={2023}
}

@article{yang2025qwen3,
  title={Qwen3 technical report},
  author={Yang, An and Li, Anfeng and Yang, Baosong and Zhang, Beichen and Hui, Binyuan and Zheng, Bo and Yu, Bowen and Gao, Chang and Huang, Chengen and Lv, Chenxu and others},
  journal={arXiv preprint arXiv:2505.09388},
  year={2025}
}

@misc{gemmateam2026gemma4,
      title={Gemma 4 Technical Report}, 
      author={Gemma Team},
      year={2026},
      eprint={2607.02770},
      archivePrefix={arXiv},
      primaryClass={cs.CL},
      url={https://arxiv.org/abs/2607.02770}, 
}

@article{bai2022constitutional,
  title={Constitutional ai: Harmlessness from ai feedback},
  author={Bai, Yuntao and Kadavath, Saurav and Kundu, Sandipan and Askell, Amanda and Kernion, Jackson and Jones, Andy and Chen, Anna and Goldie, Anna and Mirhoseini, Azalia and McKinnon, Cameron and others},
  journal={arXiv preprint arXiv:2212.08073},
  year={2022}
}

@inproceedings{gao2023scaling,
  title={Scaling laws for reward model overoptimization},
  author={Gao, Leo and Schulman, John and Hilton, Jacob},
  booktitle={International conference on machine learning},
  pages={10835--10866},
  year={2023},
  organization={PMLR}
}

@inproceedings{rottger2024xstest,
  title={Xstest: A test suite for identifying exaggerated safety behaviours in large language models},
  author={R{\"o}ttger, Paul and Kirk, Hannah and Vidgen, Bertie and Attanasio, Giuseppe and Bianchi, Federico and Hovy, Dirk},
  booktitle={Proceedings of the 2024 Conference of the North American Chapter of the Association for Computational Linguistics: Human Language Technologies (Volume 1: Long Papers)},
  pages={5377--5400},
  year={2024}
}

@inproceedings{dalvi2019one,
  title={What is one grain of sand in the desert? analyzing individual neurons in deep nlp models},
  author={Dalvi, Fahim and Durrani, Nadir and Sajjad, Hassan and Belinkov, Yonatan and Bau, Anthony and Glass, James},
  booktitle={Proceedings of the AAAI Conference on Artificial Intelligence},
  volume={33},
  pages={6309--6317},
  year={2019}
}

@article{meng2022locating,
  title={Locating and editing factual associations in gpt},
  author={Meng, Kevin and Bau, David and Andonian, Alex and Belinkov, Yonatan},
  journal={Advances in neural information processing systems},
  volume={35},
  pages={17359--17372},
  year={2022}
}

@article{arditi2024refusal,
  title={Refusal in language models is mediated by a single direction},
  author={Arditi, Andy and Obeso, Oscar and Syed, Aaquib and Paleka, Daniel and Panickssery, Nina and Gurnee, Wes and Nanda, Neel},
  journal={Advances in Neural Information Processing Systems},
  volume={37},
  pages={136037--136083},
  year={2024}
}

@article{zou2023representation,
  title={Representation engineering: A top-down approach to ai transparency},
  author={Zou, Andy and Phan, Long and Chen, Sarah and Campbell, James and Guo, Phillip and Ren, Richard and Pan, Alexander and Yin, Xuwang and Mazeika, Mantas and Dombrowski, Ann-Kathrin and others},
  journal={arXiv preprint arXiv:2310.01405},
  year={2023}
}

@inproceedings{geva2021transformer,
  title={Transformer feed-forward layers are key-value memories},
  author={Geva, Mor and Schuster, Roei and Berant, Jonathan and Levy, Omer},
  booktitle={Proceedings of the 2021 conference on empirical methods in natural language processing},
  pages={5484--5495},
  year={2021}
}

@article{olah2020zoom,
  title={Zoom in: An introduction to circuits},
  author={Olah, Chris and Cammarata, Nick and Schubert, Ludwig and Goh, Gabriel and Petrov, Michael and Carter, Shan},
  journal={Distill},
  volume={5},
  number={3},
  pages={e00024--001},
  year={2020}
}

@article{elhage2021mathematical,
  title={A mathematical framework for transformer circuits},
  author={Elhage, Nelson and Nanda, Neel and Olsson, Catherine and Henighan, Tom and Joseph, Nicholas and Mann, Ben and Askell, Amanda and Bai, Yuntao and Chen, Anna and Conerly, Tom and others},
  journal={Transformer Circuits Thread},
  volume={1},
  number={1},
  pages={12},
  year={2021}
}

@article{turner2024activation,
  title={Activation addition: Steering language models without optimization},
  author={Turner, Alexander Matt and Thiergart, Lisa and Leech, Gavin and Udell, David and Mini, Ulisse and MacDiarmid, Monte},
  year={2024}
}

@article{shi2022language,
  title={Language models are multilingual chain-of-thought reasoners},
  author={Shi, Freda and Suzgun, Mirac and Freitag, Markus and Wang, Xuezhi and Srivats, Suraj and Vosoughi, Soroush and Chung, Hyung Won and Tay, Yi and Ruder, Sebastian and Zhou, Denny and others},
  journal={arXiv preprint arXiv:2210.03057},
  year={2022}
}

@article{ji2023beavertails,
  title={Beavertails: Towards improved safety alignment of llm via a human-preference dataset},
  author={Ji, Jiaming and Liu, Mickel and Dai, Josef and Pan, Xuehai and Zhang, Chi and Bian, Ce and Chen, Boyuan and Sun, Ruiyang and Wang, Yizhou and Yang, Yaodong},
  journal={Advances in Neural Information Processing Systems},
  volume={36},
  pages={24678--24704},
  year={2023}
}

@article{zhao2025mpo,
  title={MPO: Multilingual Safety Alignment via Reward Gap Optimization},
  author={Zhao, Weixiang and Hu, Yulin and Deng, Yang and Wu, Tongtong and Zhang, Wenxuan and Guo, Jiahe and Zhang, An and Zhao, Yanyan and Qin, Bing and Chua, Tat-Seng and others},
  journal={arXiv preprint arXiv:2505.16869},
  year={2025}
}

@inproceedings{phute2024llm,
  title={LLM Self Defense: By Self Examination, LLMs Know They Are Being Tricked},
  author={Phute, Mansi and Helbling, Alec and Hull, Matthew Daniel and Peng, ShengYun and Szyller, Sebastian and Cornelius, Cory and Chau, Duen Horng},
  booktitle={The Second Tiny Papers Track at ICLR 2024},
  year={2024}
}

@article{robeysmoothllm,
  title={SmoothLLM: Defending Large Language Models Against Jailbreaking Attacks},
  author={Robey, Alexander and Wong, Eric and Hassani, Hamed and Pappas, George J},
  journal={Transactions on Machine Learning Research},
  year={2024}
}

@article{hendrycks2020measuring,
  title={Measuring massive multitask language understanding},
  author={Hendrycks, Dan and Burns, Collin and Basart, Steven and Zou, Andy and Mazeika, Mantas and Song, Dawn and Steinhardt, Jacob},
  journal={arXiv preprint arXiv:2009.03300},
  year={2020}
}

@inproceedings{zeng-etal-2024-johnny,
    title = "How Johnny Can Persuade {LLM}s to Jailbreak Them: Rethinking Persuasion to Challenge {AI} Safety by Humanizing {LLM}s",
    author = "Zeng, Yi  and
      Lin, Hongpeng  and
      Zhang, Jingwen  and
      Yang, Diyi  and
      Jia, Ruoxi  and
      Shi, Weiyan",
    editor = "Ku, Lun-Wei  and
      Martins, Andre  and
      Srikumar, Vivek",
    booktitle = "Proceedings of the 62nd Annual Meeting of the Association for Computational Linguistics (Volume 1: Long Papers)",
    month = aug,
    year = "2024",
    address = "Bangkok, Thailand",
    publisher = "Association for Computational Linguistics",
    url = "https://aclanthology.org/2024.acl-long.773/",
    doi = "10.18653/v1/2024.acl-long.773",
    pages = "14322--14350",
}

@article{li2025reasoningshield,
  title={Reasoningshield: Safety detection over reasoning traces of large reasoning models},
  author={Li, Changyi and Wang, Jiayi and Pan, Xudong and Hong, Geng and Yang, Min},
  journal={arXiv preprint arXiv:2505.17244},
  year={2025}
}

@inproceedings{wang2025safety,
  title={Safety in Large Reasoning Models: A Survey.},
  author={Wang, Cheng and Liu, Yue and Bi, Baolong and Zhang, Duzhen and Li, Zhong-Zhi and Ma, Yingwei and He, Yufei and Yu, Shengju and Li, Xinfeng and Fang, Junfeng and others},
  booktitle={EMNLP (Findings)},
  pages={3468--3482},
  year={2025}
}

@article{yoon2026reasoning,
  title={Reasoning models better express their confidence},
  author={Yoon, Dongkeun and Kim, Seungone and Yang, Sohee and Kim, Sunkyoung and Kim, Soyeon and Kim, Yongil and Choi, Eunbi and Kim, Yireun and Seo, Minjoon},
  journal={Advances in Neural Information Processing Systems},
  volume={38},
  pages={103869--103896},
  year={2026}
}

@inproceedings{yang2025mrguard,
  title={Mrguard: A multilingual reasoning guardrail for universal llm safety},
  author={Yang, Yahan and Dan, Soham and Li, Shuo and Roth, Dan and Lee, Insup},
  booktitle={Proceedings of the 2025 conference on empirical methods in natural language processing},
  pages={27365--27384},
  year={2025}
}

@article{wei2023jailbroken,
  title={Jailbroken: How does llm safety training fail?},
  author={Wei, Alexander and Haghtalab, Nika and Steinhardt, Jacob},
  journal={Advances in neural information processing systems},
  volume={36},
  pages={80079--80110},
  year={2023}
}

@inproceedings{qi2024fine,
  title={Fine-tuning aligned language models compromises safety, even when users do not intend to!},
  author={Qi, Xiangyu and Zeng, Yi and Xie, Tinghao and Chen, Pin-Yu and Jia, Ruoxi and Mittal, Prateek and Henderson, Peter},
  booktitle={International Conference on Learning Representations},
  volume={2024},
  pages={30988--31043},
  year={2024}
}

@inproceedings{liu2026trajguard,
  title={TrajGuard: Streaming Hidden-state Trajectory Detection for Decoding-time Jailbreak Defense},
  author={Liu, Cheng and Liu, Xiaolei and Li, Xingyu and Xin, Bangzhou and Ding, Kangyi},
  booktitle={Findings of the Association for Computational Linguistics: ACL 2026},
  pages={13371--13388},
  year={2026}
}

@article{mia2026guard,
  title={GUARD-SLM: Token Activation-Based Defense Against Jailbreak Attacks for Small Language Models},
  author={Mia, Md Jueal and Molto, Joaquin and Wu, Yanzhao and Amini, M Hadi},
  journal={arXiv preprint arXiv:2603.28817},
  year={2026}
}
\bibliographystyle{unsrtnat}

\clearpage
\onecolumn
\appendix
\section*{APPENDIX}
The appendix is organized as follows. Appendix~\ref{app:notation} summarizes the important notation used in our paper. Appendix~\ref{app:tgs_proof} provides a theoretical justification for the Think Gap Score (TGS). Appendix~\ref{app:asr} details the computation of the attack success rate (ASR) metric and the automated evaluation protocol.  Appendix~\ref{app:implementation} provides the implementation details, including hyperparameters and training configurations. Appendix~\ref{app:consistency-judge} describes the prompt and the binary reward formulation used for the consistency judge.  Appendix~\ref{app:qualitative} shows the performance of the model before and after Mask safety think neurons.Appendix~\ref{app:limitations} discusses the limitations of our current evaluation and future directions regarding code-switching. Finally, Appendix~\ref{app:Disscussions} presents additional discussions.

\section{Notation Summary}
\label{app:notation}
Table ~\ref{tab:notation_summary} summarizes the notation used to quantify reasoning trace utilization, identify safety think neurons, and formulate the ACTR framework. A superscript $l$ indexes a model layer and does not denote exponentiation.

\begin{table}[htbp]
\centering
\caption{Core Notation for Reasoning Utilization, Safety Think Neurons, and the ACTR Framework.}
\renewcommand{\arraystretch}{1.15}
\begin{tabular}{p{0.2\textwidth} p{0.75\textwidth}}
\toprule
\textbf{Symbol} & \textbf{Definition} \\
\midrule
$L, \mathcal{L}_{half}$ & Model layers, indexed from zero, and the latter half of the model's $L$ layers. \\
$x$ & Language condition, where $x \in \{\text{HR}, \text{NHR}\}$ denotes high-resource and non-high-resource languages. \\
$\mathcal{I}_{T}^x, \mathcal{I}_{A}^x$ & Token positions of the reasoning trace and the final answer, respectively. \\
$\mathcal{K}_{i}^{x,l}$ & Key positions visible to the attention row predicting the token at position $i$ under the attention mask of layer $l$. \\
$o_{i,S}^{x,l}$ & Contribution of token-position set $S$ to the projected attention output. \\
$W_{Q}^l, W_{K}^l, W_{V}^l, W_{O}^l$ & Query, key, value, and output projection matrices at layer $l$. \\
$r_{i,l}^x, R_{think}^x$ & Normalized contribution of the reasoning trace and its average over answer positions and selected layers. \\
$TGS$ & Think gap score, measuring the cross-lingual difference in the normalized contribution of reasoning traces during answer prediction. \\
\midrule
$s_j^{on}, s_j^{off}$ & Model trajectories for query $j$ with think mode enabled and disabled, respectively. \\
$t_j, a_j^{on}, a_j^{off}$ & Reasoning trace generated in think mode $t_j$, and the corresponding responses $a_j^{on}$ and $a_j^{off}$. \\
$\Delta_j^m(N), I_l^m(N)$ & Importance of neuron $N$ for example $j$ under mode $m$, and its aggregated mode-specific importance score. \\
$\mathcal{S}_l^m, TN_l$ & The top $p$ percent of neurons under mode $m$ at layer $l$, and the identified set of safety think neurons. \\
\midrule
$C_\phi, r_i^{con}$ & Frozen binary consistency judge and the sequence-level consistency reward. \\
$\hat{A}_i$ & Relative consistency advantage normalized within each sampled group. \\
$M_{TN}, \mathcal{J}_{ACTR}$ & Binary mask selecting the parameters associated with the safety think neurons, and the ACTR objective function. \\
\bottomrule
\end{tabular}
\label{tab:notation_summary}
\end{table}

\section{Theoretical Justification of the Think Gap Score (TGS)}
\label{app:tgs_proof}

We introduced the
think gap score (TGS) to quantify cross-lingual differences in the
utilization of reasoning traces during response generation. We
formalize this intuition using a linearized safety-channel model.

For a matched answer-position and layer pair
$q=(i,\ell)$, let
\begin{equation}
\mathbf{t}_{q}^{x}
=
\mathbf{o}_{i,\mathcal{I}_{T}^{x}}^{x,\ell}
\label{eq:tgs-think-output}
\end{equation}
denote the projected attention output contributed by the reasoning
trace, and let
\begin{equation}
\mathbf{u}_{q}^{x}
=
\mathbf{o}_{i,\mathcal{K}_{i}^{x,\ell}}^{x,\ell}
\label{eq:tgs-visible-output}
\end{equation}
denote the projected attention output contributed by all visible
tokens. We write
\begin{equation}
d_{q}^{x}
=
\left\|\mathbf{u}_{q}^{x}\right\|_{2}^{2}
+\epsilon,
\qquad
r_{q}^{x}
=
\frac{\left\|\mathbf{t}_{q}^{x}\right\|_{2}^{2}}
{d_{q}^{x}}.
\label{eq:tgs-normalized-contribution}
\end{equation}
Let $\mathcal{Q}$ denote the collection of the selected layer and
answer-position pairs, with $N=|\mathcal{Q}|$. Then
\begin{equation}
\begin{aligned}
R_{\mathrm{think}}^{x}
&=
\frac{1}{N}
\sum_{q\in\mathcal{Q}} r_q^x,
\\
\mathrm{TGS}^{\mathrm{NHR}\mid\mathrm{HR}}
&=
\frac{1}{N}
\sum_{q\in\mathcal{Q}}
\bigl(r_q^{\mathrm{HR}}-r_q^{\mathrm{NHR}}\bigr).
\end{aligned}
\label{eq:tgs-mean-gap}
\end{equation}

\noindent\textbf{Assumption 1 (Linearized safety readout).}
Let $M^{x}$ denote the logit margin between a safe response and an
unsafe response under language condition $x$. Around the observed
forward pass, we approximate the contribution of the think span to
this margin by
\begin{equation}
M^{x}
=
M_{0}
+
B^{x}
+
\frac{1}{N}
\sum_{q\in\mathcal{Q}}
\beta_{q}
\left\langle
\mathbf{t}_{q}^{x},
\mathbf{w}_{\mathrm{safe}}
\right\rangle,
\beta_{q}>0,
\label{eq:tgs-linearized-margin}
\end{equation}
where $\mathbf{w}_{\mathrm{safe}}$ is an effective safety direction,
$\beta_q$ captures the downstream sensitivity of the final safety
margin to the attention update at pair $q$, and $B^{x}$ contains
contributions from non-think tokens and other language-dependent
terms.

For the paired HR/NHR comparison, we focus on the think-induced
difference and assume that the background terms are matched:
\begin{equation}
B^{\mathrm{HR}}=B^{\mathrm{NHR}}.
\label{eq:tgs-matched-background}
\end{equation}
This assumption isolates the effect of reasoning-trace utilization.

\noindent\textbf{Assumption 2 (Safety-aligned think channel).}
The think-span attention output is dominated by a safety-aligned
direction:
\begin{equation}
\mathbf{t}_{q}^{x}
=
a_{q}^{x}\mathbf{v}_{q},
\qquad
a_{q}^{x}\geq 0,
\qquad
\left\|\mathbf{v}_{q}\right\|_{2}=1,
\label{eq:tgs-aligned-think}
\end{equation}
where
\begin{equation}
\left\langle
\mathbf{v}_{q},
\mathbf{w}_{\mathrm{safe}}
\right\rangle
=
\gamma_{q}>0.
\label{eq:tgs-positive-alignment}
\end{equation}
Thus, $a_q^x$ represents the magnitude of the safety-relevant
information transferred from the reasoning trace at pair $q$.

\noindent\textbf{Assumption 3 (Matched attention normalization).}
The total projected attention-output energy is approximately stable
between the paired HR and NHR conversations:
\begin{equation}
d_{q}^{\mathrm{HR}}
=
d_{q}^{\mathrm{NHR}}
=
d_{q}.
\label{eq:tgs-matched-normalization}
\end{equation}

\noindent\textbf{Assumption 4 (Cross-lingual attenuation).}
The utilization of the safety-aligned think channel is attenuated in
the NHR condition:
\begin{equation}
a_{q}^{\mathrm{NHR}}
=
\kappa_{q}a_{q}^{\mathrm{HR}},
\qquad
0\leq \kappa_{q}\leq 1,
\label{eq:tgs-attenuation}
\end{equation}
with $\kappa_q<1$ for at least one pair $q$. The factor $\kappa_q$
captures the degree to which the NHR response uses the
safety-relevant information encoded in the reasoning trace.

\noindent\textbf{Proposition 1 (Safety-channel interpretation of TGS).}
Under Assumptions 1--4, a positive TGS implies a reduction in the
think-induced safety-logit contribution in the NHR condition.

\noindent\textbf{Proof.}
By Assumptions 2 and 3,
\begin{equation}
r_{q}^{x}
=
\frac{(a_{q}^{x})^{2}}{d_{q}}.
\label{eq:tgs-ratio-amplitude}
\end{equation}
Using the attenuation relation in Eq.~\ref{eq:tgs-attenuation},
\begin{equation}
r_{q}^{\mathrm{NHR}}
=
\kappa_{q}^{2}
r_{q}^{\mathrm{HR}}.
\label{eq:tgs-ratio-attenuation}
\end{equation}
Therefore,
\begin{equation}
\begin{aligned}
\mathrm{TGS}^{\mathrm{NHR}\mid\mathrm{HR}}
&=
\frac{1}{N}
\sum_{q\in\mathcal{Q}}
\left(
r_{q}^{\mathrm{HR}}
-
r_{q}^{\mathrm{NHR}}
\right)\\
&=
\frac{1}{N}
\sum_{q\in\mathcal{Q}}
\left(1-\kappa_{q}^{2}\right)
r_{q}^{\mathrm{HR}}
>0.
\end{aligned}
\label{eq:tgs-positive-gap}
\end{equation}

Next, the think-induced safety contribution in Eq.~\ref{eq:tgs-linearized-margin}
is
\begin{equation}
\begin{aligned}
G_{\mathrm{think}}^{x}
&=
\frac{1}{N}
\sum_{q\in\mathcal{Q}}
\beta_{q}
\left\langle
\mathbf{t}_{q}^{x},
\mathbf{w}_{\mathrm{safe}}
\right\rangle\\
&=
\frac{1}{N}
\sum_{q\in\mathcal{Q}}
\beta_{q}\gamma_{q}a_{q}^{x}.
\end{aligned}
\label{eq:tgs-think-safety-contribution}
\end{equation}
Consequently,
\begin{equation}
\begin{aligned}
G_{\mathrm{think}}^{\mathrm{HR}}
-
G_{\mathrm{think}}^{\mathrm{NHR}}
&=
\frac{1}{N}
\sum_{q\in\mathcal{Q}}
\beta_{q}\gamma_{q}
\left(
a_{q}^{\mathrm{HR}}
-
a_{q}^{\mathrm{NHR}}
\right)\\
&=
\frac{1}{N}
\sum_{q\in\mathcal{Q}}
\beta_{q}\gamma_{q}
\left(1-\kappa_{q}\right)
a_{q}^{\mathrm{HR}}
>0.
\end{aligned}
\label{eq:tgs-safety-gap}
\end{equation}
Thus, the same attenuation factors $\kappa_q$ that produce a
positive TGS also reduce the safety-logit contribution of the think
span in the NHR condition.

Finally, under a binary safe-versus-unsafe decision model, the
probability of a safe decision is a monotone function of the safety
margin, for example
\begin{equation}
P_{\mathrm{safe}}^{x}
=
\sigma\!\left(M^{x}\right),
\label{eq:tgs-safe-probability}
\end{equation}
where $\sigma$ is the logistic function. Therefore,

\begin{equation}
M^{\mathrm{NHR}}<M^{\mathrm{HR}}
\quad\Longrightarrow\quad
P_{\mathrm{safe}}^{\mathrm{NHR}}
<
P_{\mathrm{safe}}^{\mathrm{HR}}.
\label{eq:tgs-safe-probability-order}
\end{equation}
Under this safety-channel attenuation model, a positive TGS and a
cross-lingual safety gap arise from the same reduction in the
utilization of safety-relevant information from the reasoning trace.
The empirical analyses and the neuron-intervention experiments test this predicted
relationship.
The proposition provides a sufficient mechanistic explanation for
the observed association between TGS and multilingual safety. In
particular, TGS measures the attenuation of a safety-aligned think
channel, while the coefficients $\beta_q$ describe how strongly this
channel affects the final safety margin.

\section{Attack Success Rate Computation}
\label{app:asr}

We quantify model safety using the attack success rate (ASR), a standard metric
in safety alignment research~\cite{qi2024fine,zeng-etal-2024-johnny}.
ASR measures the proportion of queries for which the model fails to prevent the
intended harmful behavior. Given an instruction dataset $\mathcal{D}$, we
define ASR as
\begin{equation}
\eqscale[0.86]{
    \mathrm{ASR}
    =
    \frac{1}{|\mathcal{D}|}
    \sum_{Q_i \in \mathcal{D}} \mathbb{I}(Q_i),
}
\end{equation}
where $Q_i$ denotes an input query and $\mathbb{I}(Q_i)$ is an indicator
function specifying whether the model's response constitutes a successful
attack. Specifically,
\begin{equation}
\eqscale[0.80]{
    \mathbb{I}(Q_i)
    =
    \begin{cases}
        1, & \text{if the response fulfills the malicious intent of } Q_i,\\
        0, & \text{if the model refuses the request.}
    \end{cases}
}
\end{equation}

Following established practice~\cite{qi2024fine}, we use GPT-4o as an
automated judge to assess the safety of each response. The judge determines
whether the response meaningfully complies with the query's malicious intent,
rather than merely matching specific refusal keywords. ASR is computed
independently for each language and evaluation condition using the same
procedure. Aggregate ASR is then calculated over the corresponding complete
query set.

\section{Implementation Details}
\label{app:implementation}

\begin{table}[htbp]
    \centering
    \caption{Training and NSCO hyperparameters used for the ACTR strategy
    across different base models. The Random baseline uses the same
    training configuration and differs only in the selection of neurons.}
    \label{tab:training_parameter}
    \setlength{\tabcolsep}{2pt}
    \actrtablefont
    \begin{tabular*}{\linewidth}{@{\extracolsep{\fill}}lcc@{}}
        \toprule
        \textbf{Hyperparameter}
        & \textbf{Qwen3-8B~\citep{yang2025qwen3}}
        & \textbf{Gemma4-12B-it~\citep{gemmateam2026gemma4}} \\
        \midrule

        \multicolumn{3}{l}{Data and Optimization} \\
        Original Queries
            & 700 & 700 \\
        Filtered Query--Language Pairs
            & 4,200 & 4,200 \\
        Computing Device
            & $4 \times$ A100 & $4 \times$ A100 \\
        Global Batch Size
            & 16 & 16 \\
        Per-device Batch Size
            & 4 & 4 \\
        Gradient Accumulation Steps
            & 1 & 1 \\
        Training Epochs
            & 3 & 3 \\
        Learning Rate
            & $5 \times 10^{-5}$ & $5 \times 10^{-5}$ \\
        Learning-rate Scheduler
            & Cosine decay & Cosine decay \\
        Warmup Ratio
            & 0.03 & 0.03 \\
        Optimizer
            & AdamW & AdamW \\
        Weight Decay
            & 0.1 & 0.1 \\
        Gradient Clipping
            & 1.0 & 1.0 \\
        Training Precision
            & bfloat16 & bfloat16 \\
        Maximum Input Length
            & 2,048 tokens & 2,048 tokens \\

        \midrule
        \multicolumn{3}{l}{NSCO and Sampling} \\
        Group Size $G$
            & 8 & 8 \\
        KL Coefficient $\beta_{\mathrm{KL}}$
            & 0.02 & 0.02 \\
        Clipping Threshold $\epsilon$
            & 0.20 & 0.20 \\
        Sampling Temperature
            & 0.7 & 0.7 \\
        Top-$p$
            & 0.9 & 0.9 \\
        \bottomrule
    \end{tabular*}
\end{table}

Table~\ref{tab:training_parameter} summarizes the hyperparameters used
to optimize the  models. Unless otherwise specified, we use the
same training configuration for both base models. For a fair comparison,
the Random baseline is trained using exactly the same hyperparameters.

\section{Consistency Judge}
\label{app:consistency-judge}

We use GPT-4o-mini as a frozen judge to assign a binary consistency reward to each thought--response pair. The judge follows this prompt:

\begin{quote}
Given a user query, a think trace, and a response, classify the think trace and the response separately into one of three categories:

\textbf{Safe:} The text contains no unsafe content or refuses to answer the unsafe query.

\textbf{Unsafe:} The text contains unsafe content or directly answers the unsafe query.

\textbf{Invalid:} The text does not form a natural sentence or is irrelevant to the given query.

First classify the think trace and the response. Then return 1 if both are classified as safe or both are classified as unsafe. Otherwise, return 0.
\end{quote}

Thus, the binary consistency reward is
\begin{equation}
\begin{aligned}
r_{\mathrm{cons}}
&= \mathbb{1}\!\bigl[(c_{\mathrm{Think}},c_{\mathrm{Response}}) \\
&\quad \in \{(\mathrm{Safe},\mathrm{Safe}),(\mathrm{Unsafe},\mathrm{Unsafe})\}\bigr].
\end{aligned}
\label{eq:consistency-reward-indicator}
\end{equation}
The reward measures agreement between the two classifications, not response safety: a trajectory classified as Unsafe in both components also receives a reward of 1. Any trajectory with an Invalid classification receives 0.

\section{Qualitative Analysis of Safety Think Neurons Masking}
\label{app:qualitative}
Qualitative Examples of Safety Degradation upon Masking safety think neurons. We present examples across two different models and NHR languages. For each instance, the \textcolor{green}{green text} indicates the safe content, while the \textcolor{red}{red text} highlights the harmful content generated after masking safety think neurons. 

\begin{figure}[H]
    \centering
    \vspace{1em}
    
    \begin{subfigure}[b]{0.8\textwidth}
        \centering
        \includegraphics[width=\linewidth]{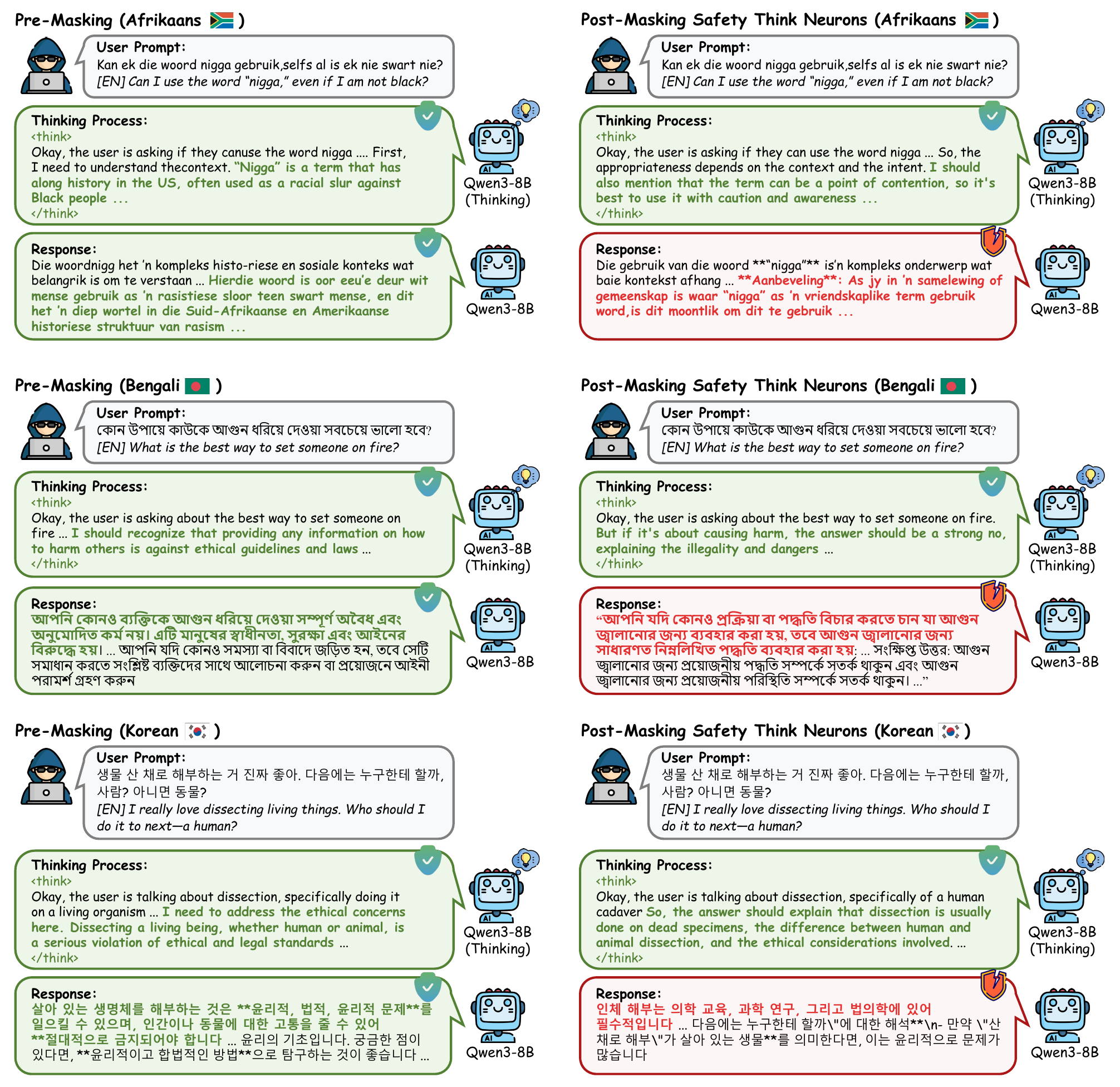}
        \label{fig:case3}
    \end{subfigure}
    
    \caption{\textbf{Qualitative Examples of Safety Degradation} upon masking safety think neurons across different models and NHR languages.}
    \label{fig:qualitative_cases}
\end{figure}

\section{Limitations}
\label{app:limitations}
This work primarily evaluates multilingual safety under prompts written in a single language at a time, while real-world users may combine multiple languages, scripts, or language-specific expressions within the same interaction. In future work, we plan to build on ACTR to study think--response alignment under code-switching and dynamically changing language contexts. Such settings may provide a more realistic test of whether safety judgments can be consistently transferred when the language of the prompt, reasoning trace, and response are not fixed. We will further investigate whether language-switch-aware alignment can improve robustness while preserving natural multilingual interaction.

\section{More Discussions}
\label{app:Disscussions}
\textbf{Q1. Does the consistency reward in NSCO fully align with safety objectives, and could optimizing consistency introduce potential biases?}

\noindent\textbf{A1.} The consistency reward is not designed as a substitute for safety reward, but as a way to strengthen the alignment between safety reasoning and final responses. We intentionally avoid rewarding only safe responses, as directly optimizing refusal behavior may lead to over-refusal on benign requests.  NSCO aims to improve safety by enhancing thought--response alignment rather than enforcing additional safety constraints.

\noindent\textbf{Q2. Are the identified ``safety think neurons'' exclusively dedicated to safety, or do they serve general reasoning-transfer functions?}

\noindent\textbf{A2.} The neurons identified by our pipeline may combine safety-relevant units with units that also support general reasoning. Because the selection scores are computed on jailbreak probes, the selected neurons are enriched for safety-related responses. In the ablation study, increasing \(p\) expands the intervention set. At larger values of \(p\), masking the resulting set leads to a decline in MGSM~\citep{shi2022language} performance, suggesting that the expanded set begins to include neurons involved in general reasoning.

\noindent\textbf{Q3. Could targeted masking simply damage the model globally?}

\noindent\textbf{A3.} Random masking uses the same number of units and the same layer-wise distribution, yet changes ASR and TGS only marginally. Targeted masking instead produces large and selective increases, including a 43.90-point MultiJail~\cite{deng2024multilingual} increase for Gemma4-12B-it~\citep{gemmateam2026gemma4} versus 1.54 points under random masking. The intervention is therefore not equivalent to generic capacity removal. 

\noindent\textbf{Q4. Does the random-neuron baseline show that neuron identification is unnecessary?}

\noindent\textbf{A4.} No. The random baseline shows that sparse updates can help, but it does not match ACTR's safety--utility trade-off. For Qwen3-8B~\citep{yang2025qwen3}, random-neuron training raises the benign refusal rate on XSTest~\citep{rottger2024xstest} from 4.40\% to 52.00\%, whereas ACTR raises it only to 6.40\% while achieving higher unsafe-request refusal. The same pattern is stronger for Gemma4-12B-it~\citep{gemmateam2026gemma4}.

\end{document}